\documentclass[journal]{IEEEtran}
\ifCLASSINFOpdf
\else
\fi
\usepackage{graphicx}
\usepackage{epstopdf}
\usepackage{subfigure}
\usepackage[justification=centering]{caption}
\usepackage{algpseudocode}
\usepackage{amsmath}
\usepackage{amssymb}
\usepackage[ruled,vlined]{algorithm2e}
\begin{document}
\title{Auction Based Clustered Federated Learning in Mobile Edge Computing System}
\author{Renhao Lu,
        Weizhe Zhang,~\IEEEmembership{Senior Member, IEEE,}
        Qiong Li, ~\IEEEmembership{ }

        Xiaoxiong Zhong,~\IEEEmembership{Member, IEEE,}
        Athanasios V. Vasilakos,~\IEEEmembership{Senior Member, IEEE}% <-this % stops a space
\thanks{Manuscript received February 27, 2021. (Corresponding author: Weizhe Zhang.)}
\thanks{Renhao Lu is with School of Computer Science and Technology, Harbin Institute of Technology, Harbin, China.}
\thanks{Weizhe Zhang is with School of Computer Science and Technology, Harbin Institute of Technology, Harbin, China, and also with Cyberspace Security Research Center, Peng Cheng Laboratory, Shenzhen, China. (Email:wzzhang@hit.edu.cn)}% <-this % stops a space
\thanks{Qiong Li is with School of Computer Science and Technology, Harbin Institute of Technology, Harbin, China.}
\thanks{Xiaoxiong Zhong is with Cyberspace Security Research Center, Peng Cheng Laboratory, Shenzhen, China.}
\thanks{Athanasios V. Vasilakos is with the School of Electrical and Data Engineering, University of Technology Sydney, Australia, with the Department of Computer Science and Technology, Fuzhou University, Fuzhou 350116, China, and with the Department of Computer Science, Electrical and Space Engineering, Lulea University of Technology, Lulea, 97187, Sweden (Email:th.vasilakos@gmail.com)
}}
\maketitle

\begin{abstract}
In recent years, mobile clients' computing ability and storage capacity have greatly improved, efficiently dealing with some applications locally. Federated learning is a promising distributed machine learning solution that uses local computing and local data to train the Artificial Intelligence (AI) model. Combining local computing and federated learning can train a powerful AI model under the premise of ensuring local data privacy while making full use of mobile clients' resources. However, the heterogeneity of local data, that is, Non-independent and identical distribution (Non-IID) and imbalance of local data size, may bring a bottleneck hindering the application of federated learning in mobile edge computing (MEC) system. Inspired by this, we propose a cluster-based clients selection method that can generate a federated virtual dataset that satisfies the global distribution to offset the impact of data heterogeneity and proved that the proposed scheme could converge to an approximate optimal solution. Based on the clustering method, we propose an auction-based clients selection scheme within each cluster that fully considers the system's energy heterogeneity and gives the Nash equilibrium solution of the proposed scheme for balance the energy consumption and improving the convergence rate. The simulation results show that our proposed selection methods and auction-based federated learning can achieve better performance with the Convolutional Neural Network model (CNN) under different data distributions.
\end{abstract}


\IEEEpeerreviewmaketitle

\begin{figure*}[htbp]
\centering
\includegraphics[width=\linewidth,scale=1.00]{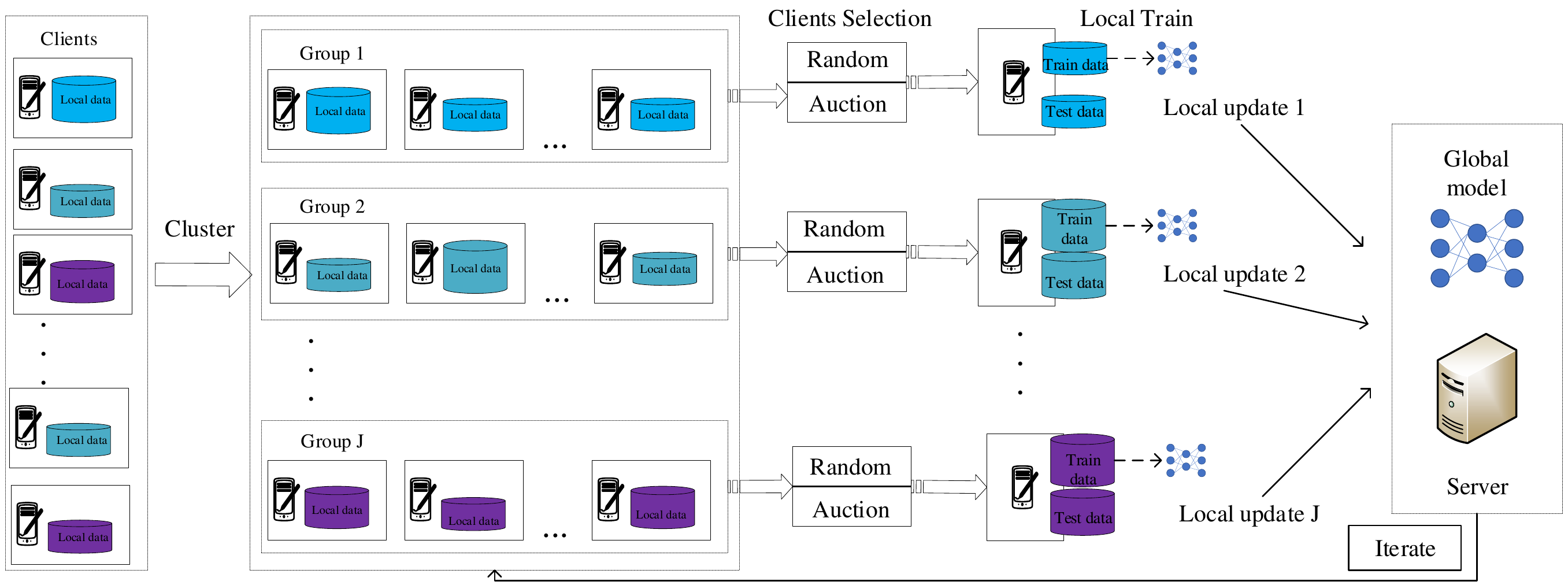}
\caption{An overview of auction based clustered federated learning.}
\label{fig1}
\end{figure*}

\section{Introduction}

\IEEEPARstart{M}{achine} learning is applied to various fields, including medical care, autonomous driving, finance, etc. Massive data generated by mobile clients can effectively promote the improvement of machine learning technology. However, these data directly or indirectly reveal the privacy of users. With the increase of people’s awareness of privacy and the introduction of privacy protection laws, privacy data leakage has become one of the main bottlenecks hindering artificial intelligence development. Federated learning \cite{Mcmahan} is proposed as a promising distributed learning to alleviate the privacy leakage problem of machine learning. Unlike traditional machine learning and distributed machine learning, there is no need to centralize user data for AI model training. In the federated learning system, clients only need to transmit the model parameters or gradients trained on their local data to the aggregation server, thereby protecting data privacy.

Moreover, data heterogeneity brings new challenges to federated learning development. Wang \emph{et al.} \cite{Wang} proposed a reinforcement learning solution to solve federated learning in Non-IID scenarios. Since this scheme requires multiple rounds of reinforcement learning model training in advance for different scenarios, its generalization ability is weak. There are also some clustering solutions to solve this challenge \cite{Sattler,Khan,Jiang,SattlerF,GhoshA}. For example, Sattler \emph{et al.} \cite{Sattler} divided the clients into several groups based on the similarity of the local model and then train in each group to improve the average accuracy. The above solutions have shown specific effects in dealing with Non-IID scenarios, but they ignore the imbalance of local data. That is, the size of data generated by different clients is inconsistent. Cai \emph{et al.}\cite{DBLP:conf/icc/CaiLZY20} proposes a scheme that using dynamic samples to solve the problem of data imbalance without considering the impact of Non-IID.

Also, especially in wireless edge networks, the energy consumption of mobile clients is usually limited. How to balance the clients' energy consumption in the system is another core challenge for developing federated learning in the mobile edge computing system. Clients are required to perform calculations, making mobile clients unwilling to participate in federated learning due to limited energy. \cite{khan2020federated,le2020auction} design incentive mechanism to give part of the benefits to participating clients to encourage them to participate in federated learning. These schemes fully consider the resource status when selecting clients and are committed to minimizing the federated learning system's overall resource consumption but ignoring individual clients' energy consumption.

\subsection{Motivation}
Federated learning can effectively solve the problem of user data privacy leakage in traditional machine learning. Furthermore, it can make full use of the remaining computing power of idle edge clients. On the one hand, federated learning systems' data heterogeneity has become one of the main bottlenecks of federated learning development. Wang \emph{et al.}\cite{Wang} observed that the clustering scheme could speed up the convergence speed of the global model compared with randomly selecting user local models on Non-IID. They also verified the effectiveness of the clustering algorithm through experiments. However, in their experimental settings, each user has the same number of data samples, which is unrealistic in real system scenarios. The real scene is different users have different data sizes, which means the scale of data owned by edge clients is imbalanced. Therefore, in our research, we will fully consider the two aspects of data heterogeneity: Non-IID and imbalances of local data. Besides, they did not give a theoretical analysis. On the other hand, partial edge clients are selected for training in each iteration. Appropriate edge client selection can effectively improve the convergence rate of the global model. However, the energy of mobile edge clients is limited. Therefore, we proposed the energy balanced selection mechanism in this paper.
%here%
\subsection{Contribution}
Fig. 1 give an overview of auction based clustered federated learning, and the main contributions of our research are as follows:

\begin{itemize}
\item We propose a client selection scheme based on initial gradient clustering, which mainly includes the following improvements:
1) We introduce the concept of federated virtual datasets, and its goal is to transform the heterogeneity of distributed local data into solving the heterogeneity of virtual datasets.
2) To alleviate the impact of local data imbalance and ensure client clustering accuracy, we propose a sample window mechanism before clustering.
3) We give a theoretical analysis of the proposed scheme and prove that it can converge to an approximate optimal solution under the stochastic gradient descent algorithm.
\item Given the uneven resource consumption caused by randomly selecting clients in the cluster, we propose a cluster internal client selection scheme based on the auction mechanism, which fully considers the data heterogeneity and each client's remaining energy. At the same time, we give the optimal solution for clients' bidding, which satisfies the Nash equilibrium.
\item We evaluate the performance of our scheme through simulation in a variety of different Non-IID scenarios. Furthermore, we introduce the metric of energy consumption balance in the federated learning scenario for the first time. The simulation results show that our scheme shows good performance in convergence rate and energy consumption balance.
\end{itemize}

\section{RELATED WORKS}
In recent years, federated learning [1] as a special distributed machine learning approach has been widely studied by researchers. On the one hand, the original intention of federated learning is to train the AI model to ensure data privacy.  \cite{DBLP:conf/icml/BhagojiCMC19,DBLP:conf/iclr/XieHCL20,DBLP:conf/infocom/WangSZSWQ19,DBLP:conf/esorics/TolpeginTGL20,DBLP:journals/corr/abs-1911-12560} study federated learning from the perspective of protecting client's data privacy and AI model. On the other hand, different from the traditional distributed machine learning system, the federated learning system's communication environment is more complex and uncertain. Therefore, reducing communication overhead and improving communication efficiency is another core challenge of  federated learning. \cite{DBLP:conf/vcip/0003HS18,DBLP:journals/corr/KonecnyMYRSB16,DBLP:conf/dcc/LiH19,DBLP:journals/tnn/ZhuJ20} is devoted to reducing the communication cost of
federated learning or the communication rounds required for training.
Recently, the heterogeneity of federated learning systems has become the main bottleneck of its development. FL heterogeneity is divided into data heterogeneity and structural heterogeneity \cite{DBLP:conf/icca/LiFL20}.

Our research is mainly to solve the challenge of data heterogeneity in the federated learning model's training process. In terms of training data samples, unlike conventional distributed machine learning, the training data samples of federated learning are generally Non-IID. McMahan \emph{et al.}\cite{Mcmahan} proposed the Federated Averaging (FedAvg) algorithm, which is a deep network federated learning method based on iterative model averaging. They also pointed out that the FedAvg algorithm is still applicable when the data of clients is Non-IID. Li \emph{et al.} \cite{DBLP:conf/iclr/LiHYWZ20} theoretically analyzed the effectiveness of the FedAvg algorithm and show that the convergence rate of FedAvg algorithm under Non-IID is significantly worse than under IID. Zhao \emph{et al.} \cite{DBLP:journals/corr/abs-1806-00582} also verified by experiment that convolutional accuracy neural networks trained with FedAvg algorithm decreases significantly under the Non-IID setting. Besides, they proposed a data-sharing strategy, which is to improve the accuracy of model training by sharing part of the data that meets the global distribution. However, without knowing the distribution of clients, it is not easy to make all client data evenly distributed by sharing data samples. \cite{DBLP:journals/jsac/WangTSLMHC19} analyzed the convergence bounds of gradient descent algorithm in federated learning system under Non-IID setting. Tian \emph{et al.} \cite{DBLP:conf/mlsys/LiSZSTS20} proposed FedProx federated learning algorithm to tackle heterogeneity in federated system. As an improved version of FedAvg, FedProx introduced a proximal term to cope with the heterogeneity of local model updating \cite{DBLP:journals/fgcs/MothukuriPPHDS21}. Wang \emph{et al.} \cite{DBLP:conf/iclr/WangYSPK20} proposed federated matched averaging (FedMA), a layer-wise federated learning algorithm designed for CNN and LSTM network architecture.

Sattler \emph{et al.} \cite{SattlerF} proposed a federated multi-task learning framework using a clustering strategy. They used the cosine similarity of the local model to cluster clients. Its purpose is to enable clients of different clusters to learn more specialized models. Ghosh \emph{et al.} \cite{GhoshA} demonstrated that each user has its learning task, and users with the same learning task can perform more efficient federated learning. Briggs \emph{et al.} \cite{DBLP:conf/ijcnn/BriggsFA20} classifies clients based on local model updates, with the goal of training specialized machine learning models over distributed datasets. However, those specialized models trained with partial local data can not make full use of the system's data. Wang et al.\cite{Wang} observed the efficiency of training a global model using clustering strategy under a Non-IID setting but did not give a theoretical analysis.

\section{CLUSTER BASED SELECTION METHOD}
\subsection{System model}

Consider a mobile edge computing scenario that consists of a cloud server $S$ and a set $NL$ of  $N$ edge clients, which act as the service provider. With the help of edge clients' data services and computing services, the server $S$ aggregates an AI model. Specifically, edge clients train the local data for $I$ rounds based on the model broadcast by the server $S$, and return the trained model parameters to the server. Server $S$ performs aggregation operations on the collected model parameters to obtain an updated model. The server and edge clients iterate the above operations until the updated model reaches the required accuracy. Besides, due to link bandwidth, timeliness of model parameters, and other reasons, the server $S$ only selects $K$ clients from $NL$ to participate in training in each iteration.  Moreover, local data samples of edge clients have heterogeneous properties, which contain two aspects: local data size and data sample distribution. Therefore, each client is divided into data of different sizes in the system model, as shown in Fig. 2.

\begin{figure}[!t]
\centering
\includegraphics[width=\linewidth,scale=1.00]{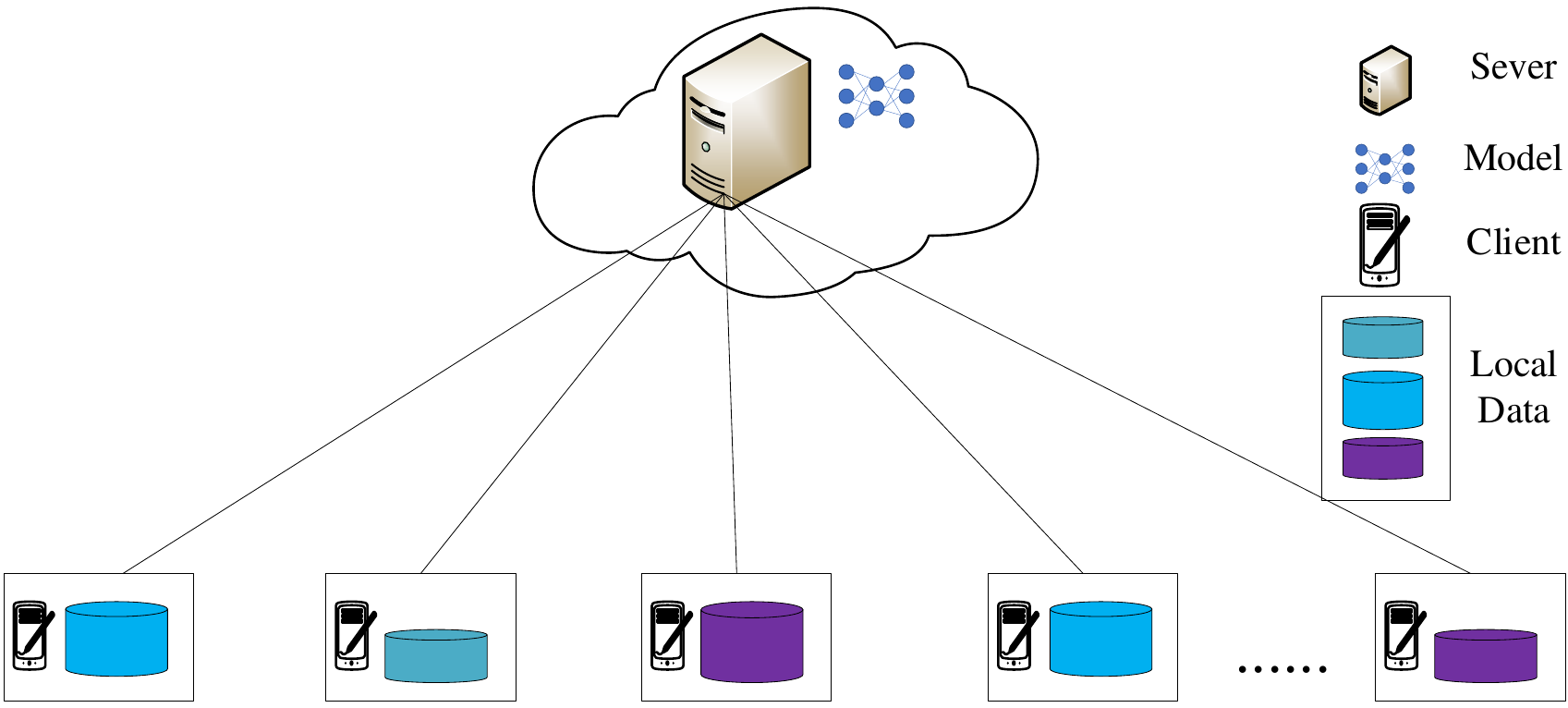}
\caption{System model.}
\label{fig2}
\end{figure}

\subsection{Federated virtual dataset}

The goal of federated learning is to jointly train a global machine learning model with massive edge clients. The training process is to gradually reduce the loss function $f(*)$, which is a distributed optimization problem:
\begin{equation}
\label{eq:eq1}
\begin{split}
    w^{*} & =\mathop{\arg\min}_{w}(f(w)) \\
	& = \mathop{\arg\min}_{w=\sum p_{k}w^{k}}(f(w^{k},\xi_{k}))
\end{split}
\end{equation}
where $w$ and $w^k$ represent the  global and local model parameters respectively, and $w^*$ is optimal global model parameters. $\xi_{k}$ and $|\xi_{k}|$ respectively denote local data and data size of client $k$. so, $p_k = \frac{|\xi_{k}|}{\sum|\xi_{k}|}$. In our research, we use stochastic gradient descent algorithm. Thus, for any client $k$ participating in training, the update process is as formula (2):
\begin{equation}
w_{t+1}^{k} = w_{t}^{k} - \eta_t \nabla f\left (w_t ,x_{t}^{k} \right )
\end{equation}
where $x^k_t \in \xi_k$, $\eta_t$ is the learning rate. Besides, let $g\left ( w_t\right )$ denote the aggregated value of local update gradient in round $t$, then,
\begin{equation}
	g\left ( w_t\right )=\sum_{k\in NL}^{NL}p_k\nabla f\left (w_t ,x_{t}^{k} \right )
\end{equation}

The large number of clients and limited communication links in the mobile edge computing network determines that not all clients can participate in each training round. Therefore, in a simple federated learning system, partial clients are selected to participate in training in each iteration, assuming that $K$ clients are selected in each round.

Let $ \xi_{t}=\bigcup_{k}^{K}x_{t}^{k}$, which is defined as a virtual datasets, then,
\begin{equation}
	g\left ( w_t,\xi_{t}\right )=\sum_{x_{t}^{k} \in \xi_{t}}p_k\nabla f\left (w_t ,x_{t}^{k} \right ).
\end{equation}

Therefore, the distributed stochastic gradient optimization algorithm of federated learning can be regarded as a traditional centralized stochastic batch gradient descent algorithm on virtual datasets $\xi_{t}$.

The above analysis is based on the situation that the selected clients only trains one local round. However, in order to alleviate the communication pressure of the system, it is general that the selected client performs $I (I \geq 2)$ local rounds training based on local datasets, that is, $I$ local rounds stochastic gradient descent algorithm or stochastic mini-batch gradient descent algorithm. So, the model parameters update rule is as follow:
\begin{equation}
	w_{t+1}=w_{t}-\eta_{t}\sum_{k=1}^{K}p_{k}\sum_{i=0}^{I}\nabla f(w^k_{t,i},x_{t,i}^{k})
\end{equation}

In our research, we only analyze the case that the learning rate $\eta_{t}$ is fixed. Furthermore, we assume: from the perspective of expectation, multi-step learning with a small learning rate is equivalent to a few steps with a large learning rate. As shown in formula (6):
\begin{equation}\label{eq:eq6}
\begin{split}
E\left [ \eta_t\sum_{k}^{K}p_k\sum_{i=0}^{I-1}\nabla f(w^{k}_{t,i},x^{k}_{t,i})\right ]\approx \\
E\left [ \theta I\eta_t\sum_{k=1}^{K}p_k\nabla f(w^{k}_{t},\xi_{k})\right ]
\end{split}
\end{equation}
where $ 0 \textless\theta\leq 1$. And, we let $\xi_{t}=\bigcup_{k}^{K}\xi_{k}$ , then,
\begin{equation}
g\left ( w_t,\xi_{t}\right )= E\left [ \sum_{k=1}^{K}p_k\nabla f(w^{k}_{t},\xi_{k})\right ]
\end{equation}
\begin{equation}
	w_{t+1}=w_{t}-\theta I\eta_tg\left ( w_t,\xi_{t}\right ).
\end{equation}
Based on this assumption, it returns to the situation of $I=1$. Thus, we give the following analyses:
\begin{enumerate}
  \item[i)] Because of local data heterogeneity, $\xi_{t}$ also has heterogeneity.
  \item[ii)] Unlike the fixity of local data distribution, the distribution of $\xi_{t}$ varies with the combination of selected clients and can be changed through the clients' selection scheme.
  \item[iii)] At this point, solving the heterogeneity problem of local data can be transformed into the heterogeneity problem of $\xi_{t}$.
  \item[iv)] As shown in Fig. 3, for any $t$, $\xi_{t}$ is consistent with the global distribution. Then, the negative impact of data heterogeneity can be alleviated.
\end{enumerate}

\begin{figure}[!t]
\centering
\includegraphics[width=\linewidth,scale=1.00]{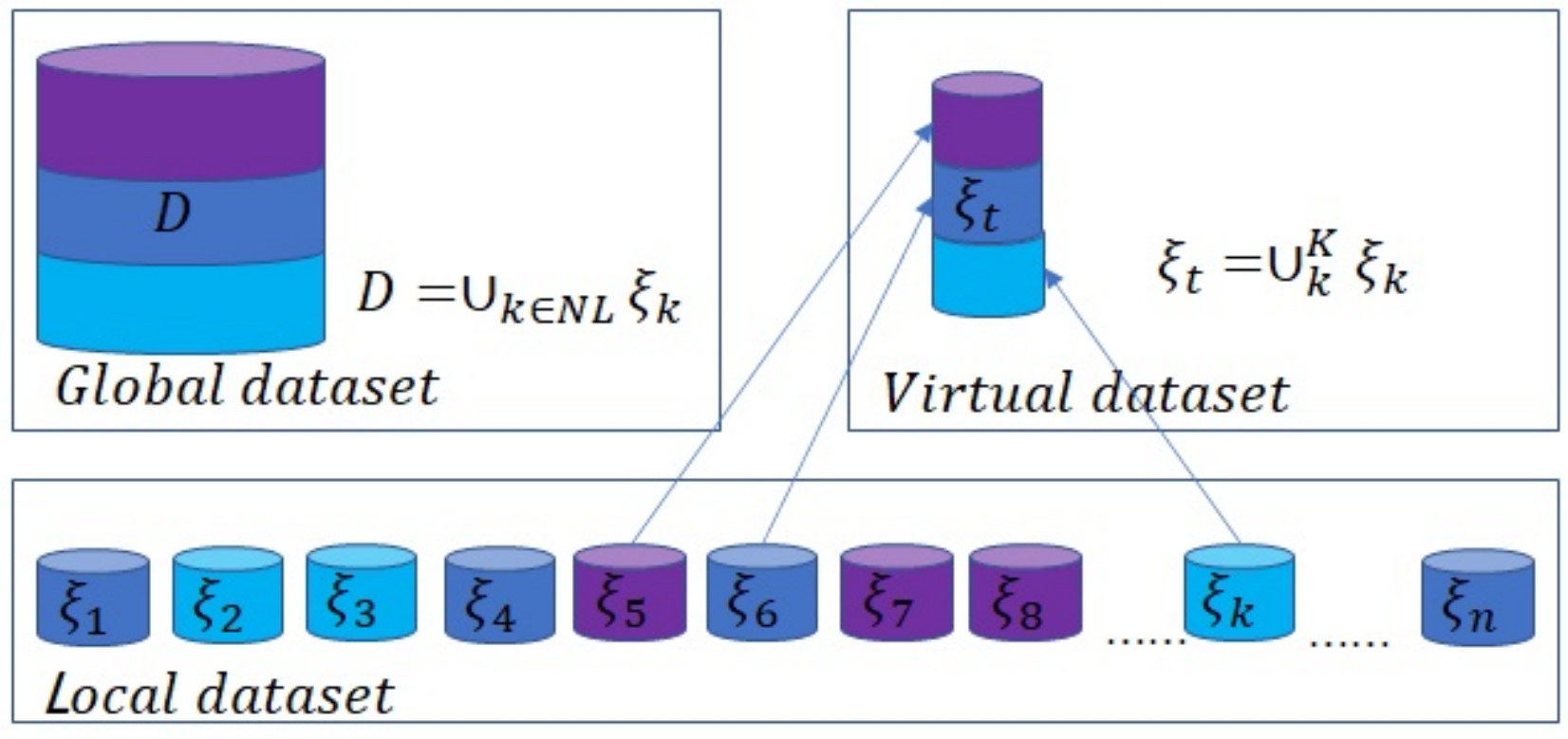}
\caption{Federated virtual dataset.}
\label{fig3}
\end{figure}

\subsection{Cluster based clients selection }
In this subsection, we propose a client selection scheme based on a clustering strategy so that the virtual dataset constructed in each round can meet the global distribution. Wang \emph{et al.}\cite{Wang} gave a client selection strategy based on initial model parameter clustering, precisely: global grouping clients based on the similarity of local models and selecting a client from each group. They also verified the effectiveness of this scheme through experiments. However, there are still two problems with t
heir strategy. Firstly, in their experimental settings, each client has the same number of data samples. In a simple federated learning system, different clients have a different number of local data samples. Secondly, after only one epoch of SGD update, the local model parameters may not reflect the local distribution.

For solving the existing problems, we propose a client selection scheme based on initial gradient clustering. Our proposed scheme mainly includes two stages: the clustering stage, training stage. In the clustering stage, there are three optimizations whose purpose is to reflect clients' distribution with more data accurately.
\begin{enumerate}
  \item Our proposed scheme uses the local gradient to represent the data distribution.
  \item We set a local sample window to limit the number of samples that the client participates in training to offset the impact of data imbalance.
  \item Before clustering, each client samples multiple times from their local data to participate in training and calculate its average gradient.
\end{enumerate}

In the training phase, we set the sample threshold to make the selected clients’ local data size at the same level, whose purpose is to lower the impact of local data imbalance. That is, randomly select a client and use its local data size as the sample threshold. After that, all clients larger than this value can participate in the selection to ensure that the virtual dataset is closer to the global distribution.

Besides, we conducted a theoretical analysis of this scheme, and we prove that the proposed scheme can converge to an approximate optimal solution with local SGD in the $\beta $ convex setting, as shown in \textbf{Theorem 1}. Please refer to the appendix A for the details of the proof.

\textbf{Theorem 1} Under the assumptions 1 to 4, which are defined in Appendix $A$, adopting clustering clients sampling strategy with fixed step size $\eta_t=\eta$ satisfies:
\begin{multline*}
E\left [ f(w_{t+1})\right ]-f(w^*)\le\\
(1-B_1)^{t-1}(f(w_1)-f(w^*)-A_1)+A_1
\end{multline*}
where $0<\eta\le \frac{\mu_E}{LM_{G}}$, $A_1=\frac{2\eta \theta IL^2M}{\mu_{E} \beta ^2}$, $B_1=\frac{\theta I \eta \mu_{E} \beta^2}{4L}.\\$

\textbf{\emph{Proof}}: see Appendix $A$ for the proof.

We also verify the effectiveness of our proposed scheme (represented as Gradients$\_$Cluster$\_$Random) through experiments. And we use Weights$\_$Cluster$\_$Random to represent the scheme proposed by \cite{Mcmahan}. In order to ensure the fairness of the evaluation, both schemes adopt the K-means clustering scheme. We also compare with the FedAvg\cite{Mcmahan} scheme (represented as FedAvg$\_$Random), which randomly selects $k$ clients for model training during each round. Furthermore, we train CNN models with Pytorch on three different datasets, MNIST, Fashion MNIST, and CIFAR-10. For data distribution settings, each client only has data for one label but random local data size. Specifically, we randomly assign a different number of samples to each client, and each client can have at least 100 data samples and a maximum of 1200 data samples.

\begin{figure}[htbp]
\begin{minipage}[t]{0.5\linewidth}
\includegraphics[width=4.2cm]{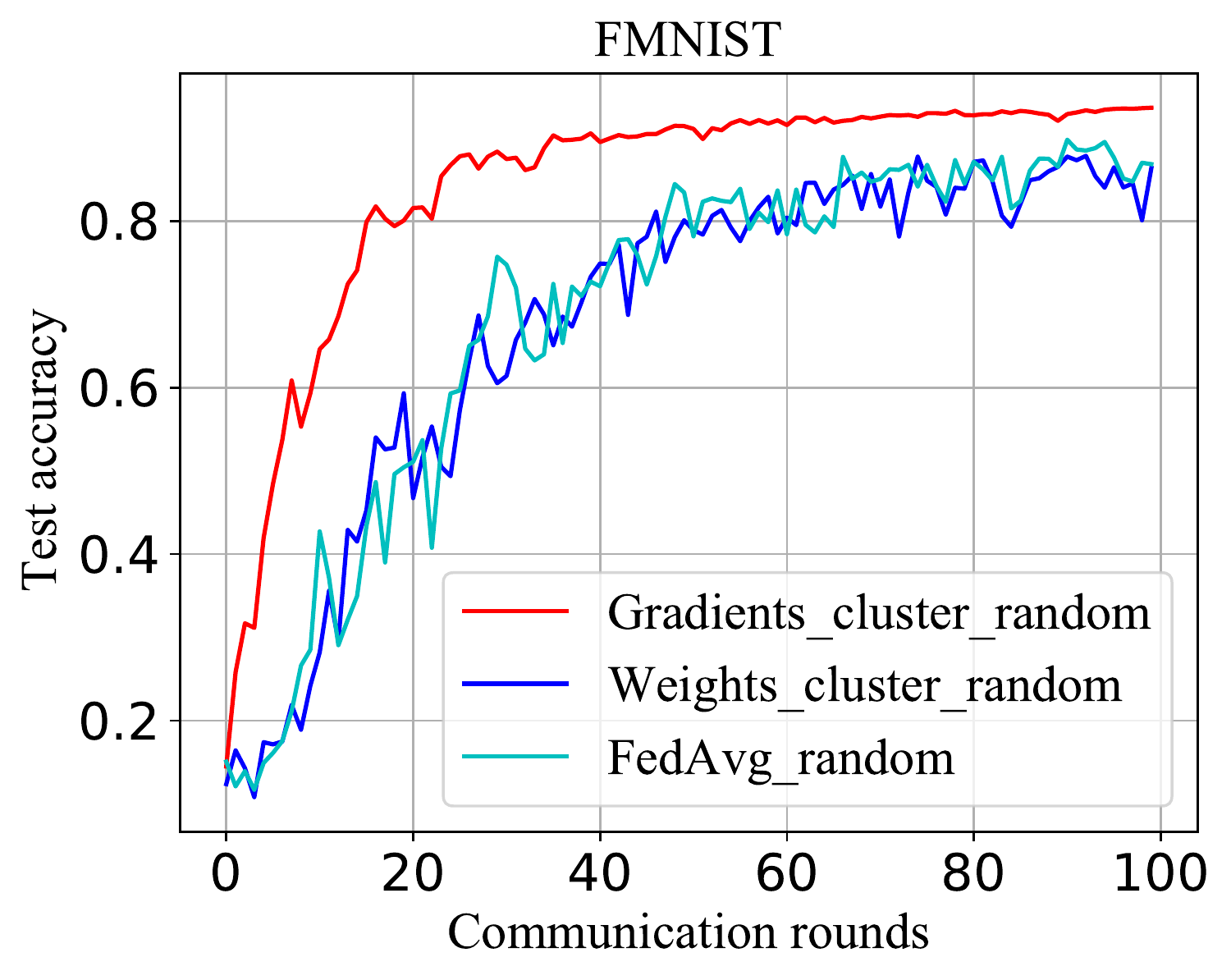}
\includegraphics[width=4.2cm]{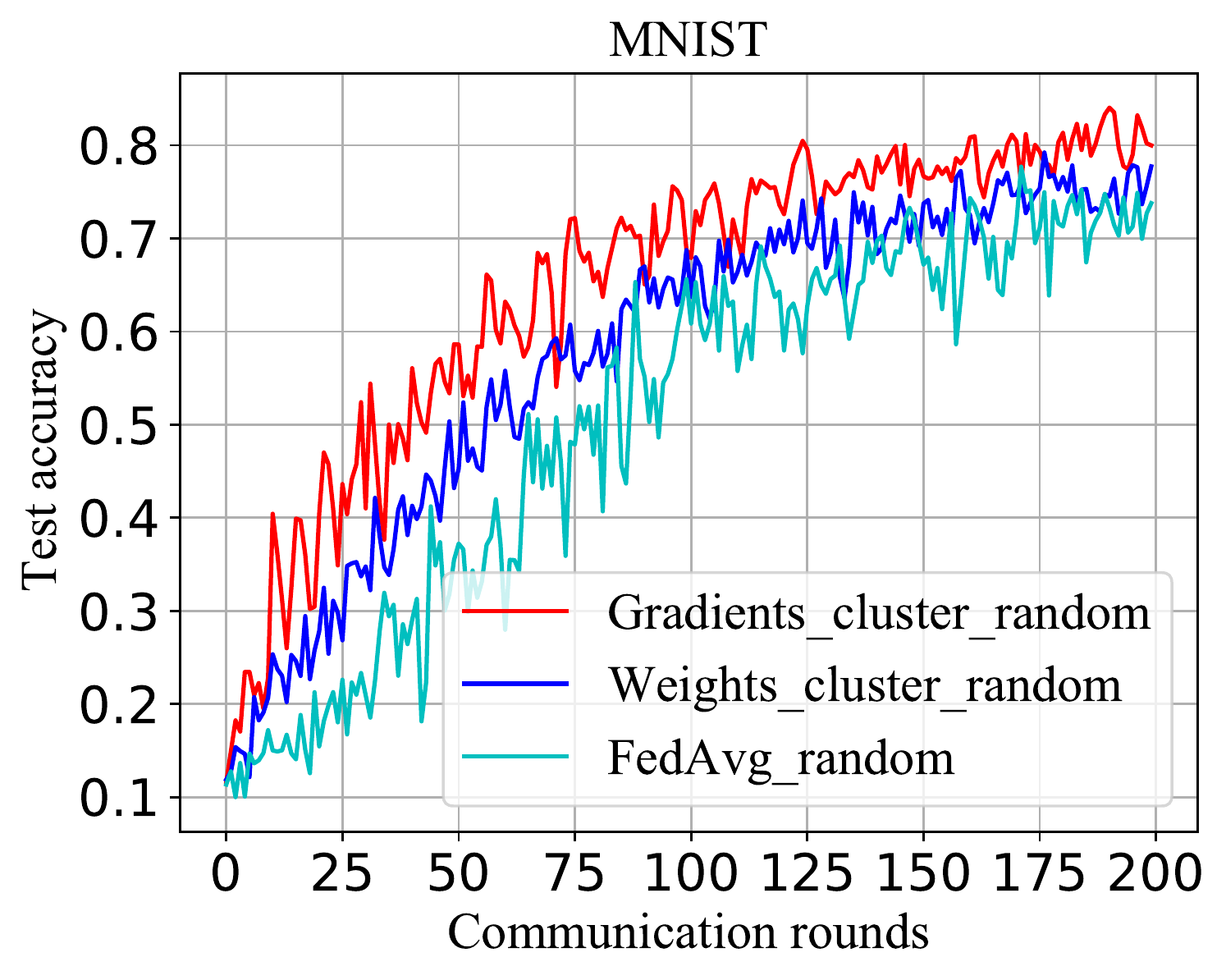}
\includegraphics[width=4.2cm]{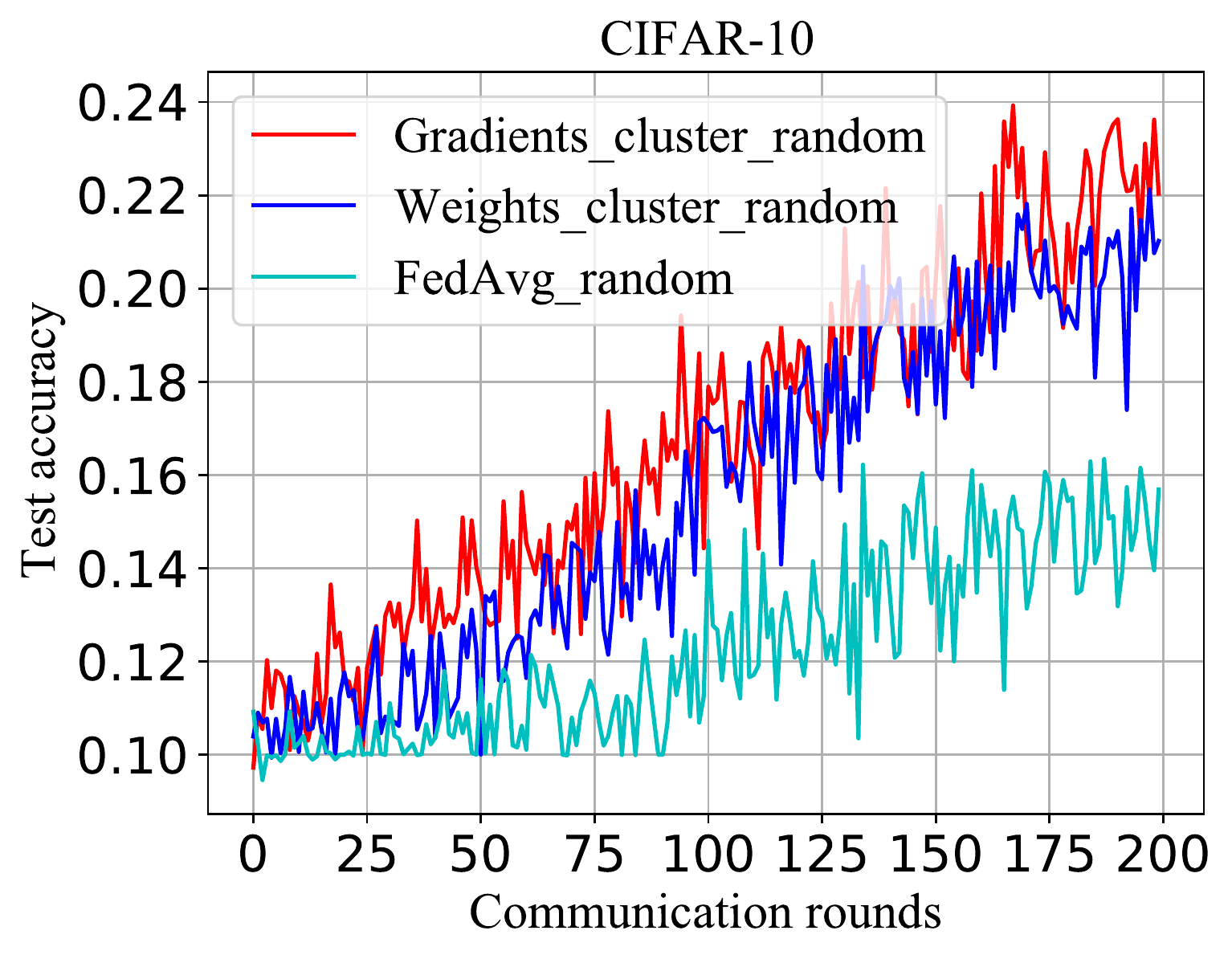}
\end{minipage}%
\begin{minipage}[t]{0.5\linewidth}
\includegraphics[width=4.3cm]{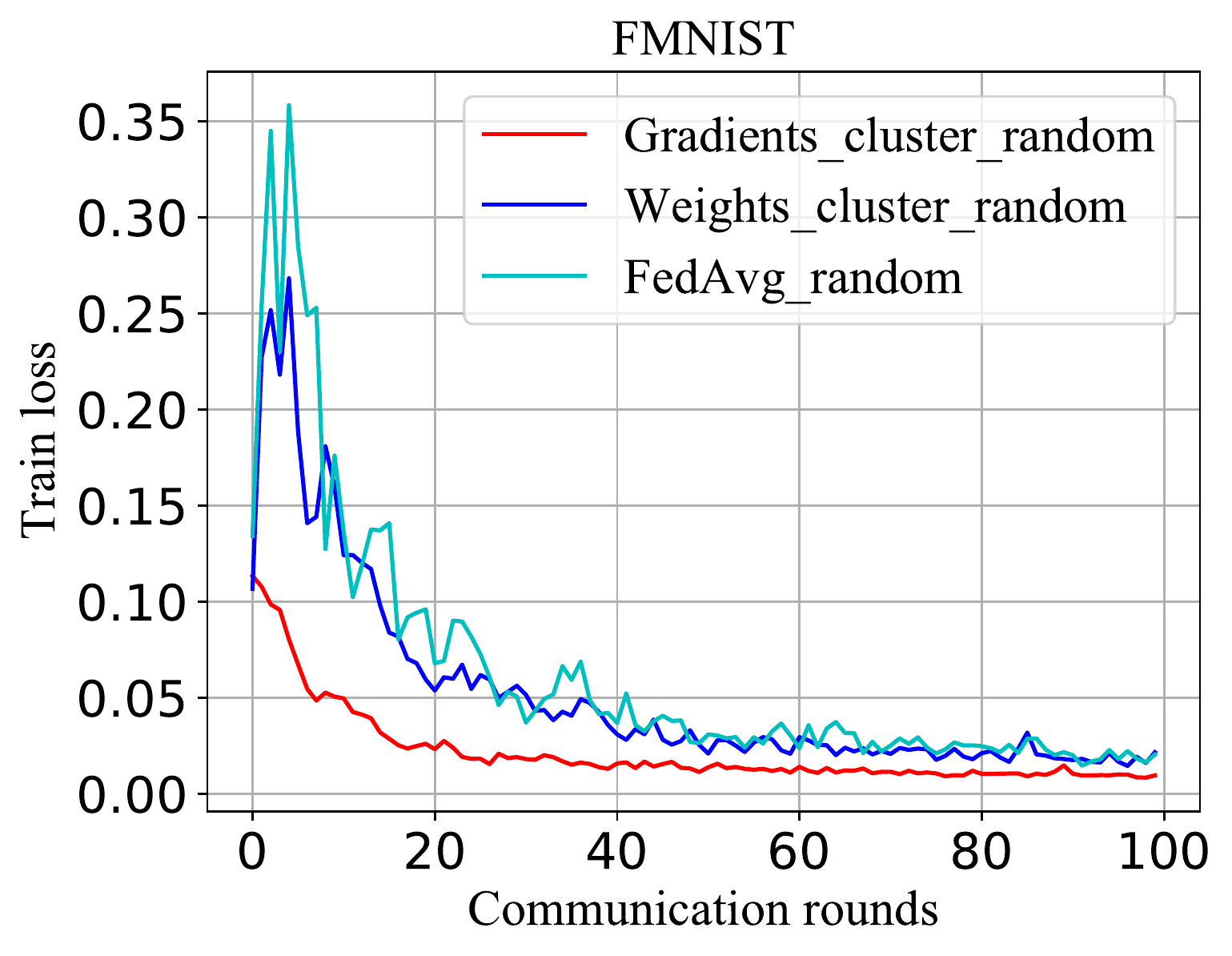}
\includegraphics[width=4.3cm]{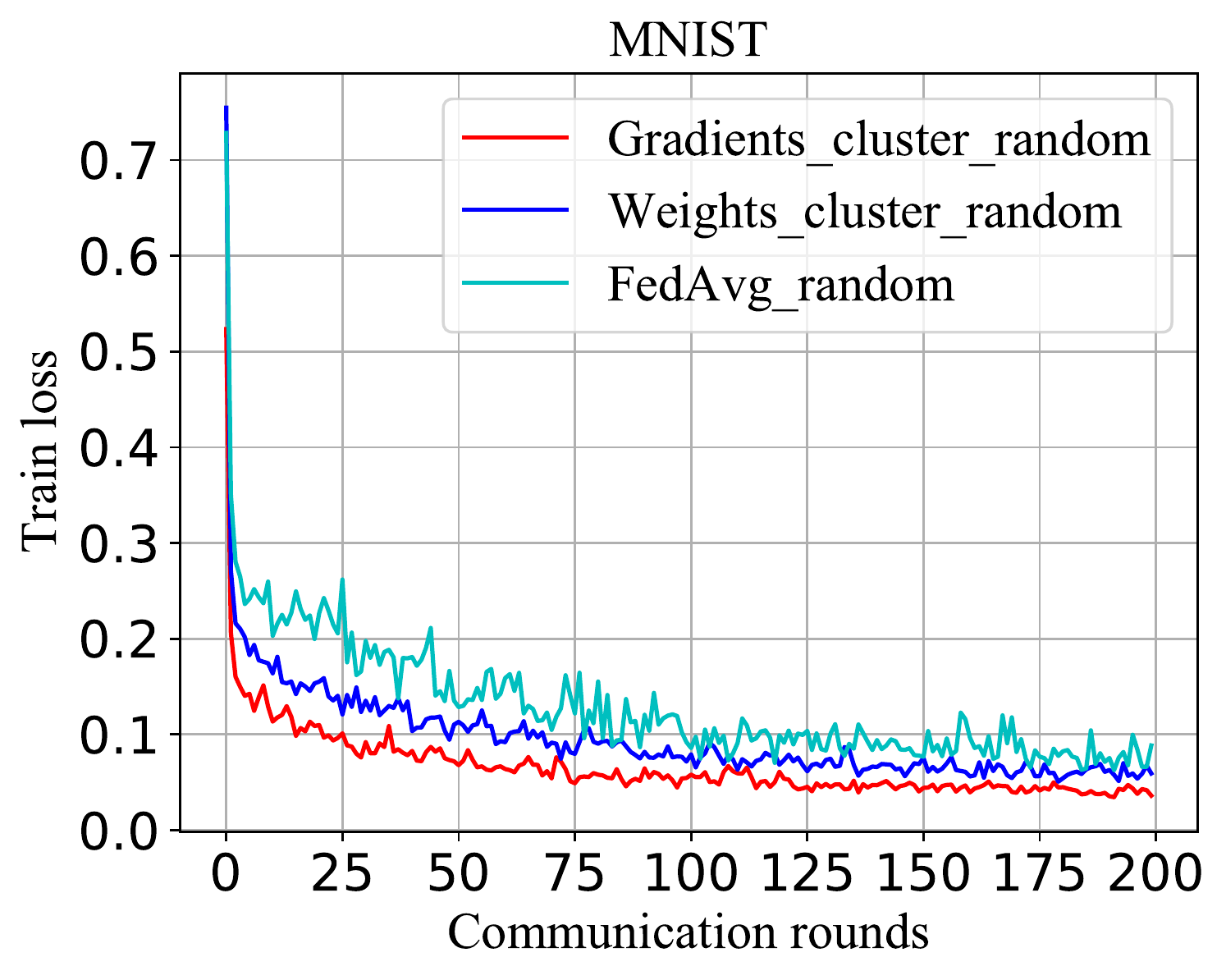}
\includegraphics[width=4.3cm]{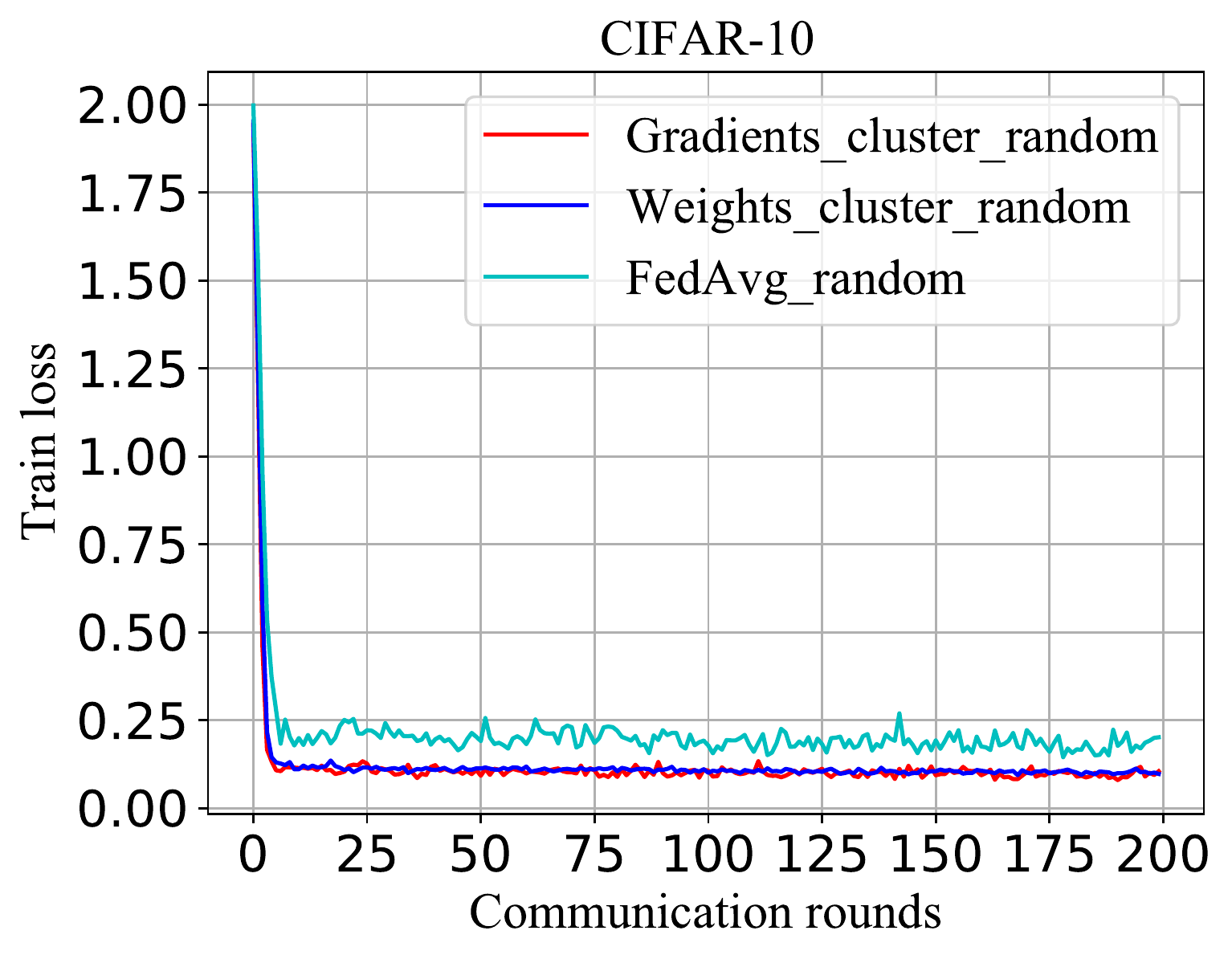}
\end{minipage}
\caption{Test accuracy and training loss v.s. communication rounds under 100 clients.}
\end{figure}

As shown in Fig. 4, the client selection schemes with clustering strategy shows better performance than FedAvg scheme. The reason is that virtual datasets constructed by the first two scheme is closer to the global distribution, and the distribution of the virtual dataset in each iteration is similar, thereby reducing the impact of data heterogeneity. Since our proposed scheme fully considers the imbalance of the data, the clustering is more accurate.
% In addition, optimized scheme adopts data truncation during training to make the distribution of virtual dataset closer to the global distribution.
Therefore, our scheme has a higher convergence rate than the other two schemes.

\section{AUCTION-BASED CLUSTERED FEDERATED LEARNING}
In section III, we experimentally verify that the clustering scheme can effectively improve global model training convergence under the Non-IID and imbalanced setting. However, in wireless edge networks, most clients are mobile clients whose energy is limited. Randomly selecting clients from each cluster for the next round of training will cause some clients' excessive consumption. Therefore, in this section, we propose an auction-based client selection scheme in each cluster. The auction mechanism is a game theory model which consists of two roles, auctioneers and bidders. As shown in Fig. 1, each group constitutes an auction system. Unlike traditional cloud computing systems, edge clients act as a bidder in this auction system, providing data services and computing services. Also, the cloud aggregation server acts as an auctioneer, and the server trains a global AI model by purchasing data services and computing services from the bidders. After bidding, there will be $K_j$ $(K_j\geq1)$ winner in cluster $j$, whose bids are the lowest within the cluster.

Our auction-based federated learning system mainly consists of four parts, energy consumption, cost function and reward model, and auction-based edge clients selection algorithm. We will describe them in detail as follows.
% \begin{figure}[!t]
% \centering
% \includegraphics[width=3.5in]{IEEEtran/Fig/Illustration of the auction mechanism.jpg}
% \caption{Illustration of the auction mechanism.}
% \label{illustion of auction based cluster}
% \end{figure}
\subsection{Energy consumption model}
In the federated learning system, each selected client $i$ trains the global model  $w_t$ on local data to obtain the local model $w_t^i$, and then all selected clients transmit the local model to the aggregation server. Therefore, in each iteration, the clients’ energy consumption consists of two parts, communication consumption, and computational consumption. The energy consumption of the edge clients can be expressed as follows:
\begin{equation}
	E_{i,t}^{sum}=E_{i,t}^{cp}+E_{i,t}^{cm}
\end{equation}
where $E_{i,t}^{sum}$ denotes the energy consumption of client $i$ as a training client in round $t$, $E_{i,t}^{cp}$ and $E_{i,t}^{cm}$ respectively represent the computation and communication consumption of the client's energy. Formula (10) represents the communication energy consumption of the client $i$.
\begin{equation}
	E_{i,t}^{cm}=E_{i,t}^{re}+E_{i,t}^{se}
\end{equation}
\noindent where $E_{i,t}^{re}$ and $E_{i,t}^{se}$ respectively represent the energy consumption of the client $i$ when receiving the global model and sending the local model. The computational energy consumption of the client $i$ is as follows:
\begin{equation}
	E_{i,t}^{cp}=(Ns_{i}\times\varrho)/100.
\end{equation}
Where $\varrho$ represents the energy consumption of each client training 100 samples and ${Ns}_i$ represents the size of the local data sample of client $i$.

\subsection{Cost function}
In our scheme, we design a cost function to determine which clients are selected for each training round. The cost function mainly depends on the client's residual energy, the number of local data samples, and the client's historical training rounds. Therefore, our cost function includes two parts, resource cost and service cost, which are represented by ${Cr}_{i,t}$ and  ${Cs}_{i,t}$, respectively.
\subsubsection{The resource cost}
The client's resource cost is determined by the residual energy of the client $i$ and the energy required by the current of training round $t$. Simultaneously, the resource cost is also dynamic, and the cost increases with the reduction of residual energy. So, the resource cost of client $i$ in round $t$ is defined as follows:
\begin{equation}
	Cr_{i,t}=\left\{\begin{matrix}
 \phi^{E_{i,t}^{res}-E_{i,t}^{cp}}&if\left ( {E_{i,t}^{res}-E_{i,t}^{cp}}>0\right ) \\
 +\infty & otherwise
\end{matrix}\right.
\end{equation}
\noindent where $0<\phi<1$, and $E_{i,t}^{res}$ represent remaining energy of the client $i$. The formula (12) also ensures that clients with sufficient energy resources have a greater probability of participating in this training round.
\subsubsection{The service cost}
The quality of service provided by clients determines the service cost. Moreover, clients' quality of service depends on the quality and quantity of their samples. Clients with more data samples can accelerate the model's convergence, which means that the clients have a better quality of service. To ensure that clients with higher service quality have a higher probability of participating in training, we set the sample size to be inversely proportional to the cost. Besides, for the global model, the more data samples involved in training can improve the model's generalization ability. Therefore, we record the historical participation rounds of clients. With the increase of clients' participation rounds, our model appropriately reduces service quality to ensure that the clients with less sample number can participate in the training. So, our service cost is calculated as follows:
\begin{equation}
	{Cs}_{i,t}=\chi \vartheta^{{Ns}_i}+\zeta (1-\log_a({co}_{i,t}+a))
\end{equation}
where ($\chi+\zeta)=1$, $(0\leq\chi$, $\zeta\geq1$, $a>1)$, ${co}_{i,t}$ represents the historical training rounds of client $i$ up to round $t$. So, our cost function is shown in formula (14).
\begin{equation}
	{c}_{i,t}=\alpha{Cs}_{i,t}+{\gamma Cr}_{i,t}
\end{equation}
% Need to change
where ($\alpha+\gamma)=1$, $(0<\alpha$, $\gamma<1)$.
\subsection{Reward model}
In this subsection, we mainly design two kinds of reward models. One is that all the benefits only belong to the clients; the other is that clients and servers share the AI model's benefits. For any client $i$, its reward in each round is expressed as $R_{i,t}^{win}$. For the first one, in the federated learning system, clients participating in the training share the global model's benefits. In line with the idea of more work, more rewards, we design a reward model suitable for federated learning, and the goal is to motivate more clients to participate in training. We set the total training round of the global training model to achieve the target accuracy rate as $Nr$, and divide the profit of the global model into each round averagely. Therefore, our reward function shows below:
\begin{equation}
	R_{i,t}^{win}=\left\{\begin{matrix}
\frac{Ns_i}{\sum_{j \in Win(t)} Ns_j}\times\frac{Rg}{Nr} & i \in Win_{t} \\
0 & otherwise
\end{matrix}\right.
\end{equation}
Where $Win_{t}$ represents the set of clients participating in the training in the round $t$, $Rg$ is the total economic income of the AI model.
%modified in the future%
For the second, the server and clients share the economic profits of the AI model according to a certain proportional relationship. In other words, we divide the profit of AI model into each round equally, and then divide each profit equally according to the number of clients participating in the training in each round. After that, the server shares this benefit with each client participating in the training in a certain proportion. We take the bid of the client as the proportion of each benefit, so the reward function of client at this time can be expressed as:
\begin{equation}
	R_{i,t}^{win} = \left\{\begin{matrix}
b_{i,t}\times\frac{Rg}{Nr} & i \in Win_{t} \\
0 & otherwise.
\end{matrix}\right.
\end{equation}
So the total return of clients $i$ is:
\begin{equation}
	{Re}_i=\sum_{t=0}^{Nr}{R_{i,t}^{win}}.
\end{equation}

\subsection{Federated learning with auction-based selection}
\subsubsection{Optimal bid}The aggregation server selects a certain percentage of clients in each cluster to participate in the training, assuming that the number of clients in cluster $j$ is $N_j$, and $K_j$ clients participate in training in each round. Therefore, our scheme abstracts the clustering-based federated learning system into an auction scenario, where the aggregation server is the auctioneer and the clients are the bidders. Furthermore, the auction scenario of federated learning has the following conditions:

\begin{itemize}
  \item [i)]
  Client bids are independent of each other, and the bids meet [0,1] uniform distribution.
  \item [ii)]
  The bid strategy is strictly monotonically increasing, and the bid strategy between clients is an asymmetrical bidding strategy.
  \item [iii)]
  The $K_j$ clients with the lowest bids win, $DL$ represents the set of winners.
  \item [iv)]
  When the bidder bids are the same, the winning client is selected according to the service cost, followed by the resource cost.
\end{itemize}

Therefore, the revenue function of the client $U_i\left ( b_{i,t},c_{i,t}\right )$  in the cluster can be expressed as follows:
\begin{equation}
	U_{i}\left ( b_{i,t},c_{i,t}\right )=\left\{\begin{matrix}
b_{i,t}-c_{i,t} & i\in N_k\\
 0&otherwise
\end{matrix}\right.
\end{equation}
where $b_{i,t}$ and $c_{i,t}$ respectively represent the price and cost of the client $i$.

\textbf{Theorem 2}
In round t, there is an optimal bid $b_{i,t}=\frac{1}{N_j-K_j+1}+\frac{N_j-K_j}{N_j-K_j+1}c_{i,t}$ for client k in cluster j, which satisfies the Nash equilibrium in the clustered federated learning system that uses the auction mechanism.

\textbf{\emph{Proof}}: The goal of using the game model is to achieve federated learning system equilibrium. For the auction mechanism, reaching the system's equilibrium state is to maximize the expected revenue of the clients participating in the auction.
\begin{equation}
\begin{split}
\max\left ( E\left ( U_i\left ( b_{i,t}-c_{i,t}\right )\right )\right )=\\
\left ( b_{i,t}-c_{i,t}\right )\prod_{j\notin N_k}P\left ( b_{i,t}<b_{i,t}\right )
\end{split}
\end{equation}
Since the client $i$ bid satisfies a uniform distribution, formula (19) can be derived as:
\begin{equation}
\begin{split}
\max\left ( E\left ( U_i\left ( b_{i,t}-c_{i,t}\right )\right )\right )=\\\left ( b_{i,t}-c_{i,t}\right )\prod_{m\notin N_k}\left (1-f_{m}^{-1}\left ( b_{i,t}\right ) \right )
\end{split}
\end{equation}
where $f_m$ is the bid strategy of client $m$, due to the symmetry of the bid strategy, so $f_m=f_i=f$, then:
\begin{equation}
\begin{split}
\max\left ( E\left ( U_i\left ( b_{i,t}-c_{i,t}\right )\right )\right )=\\
\left ( b_{i,t}-c_{i,t}\right )\left (1-f_{m}^{-1}\left ( b_{i,t}\right ) \right )^{n-K_j}.
\end{split}
\end{equation}
The first-order optimal auction condition of equation (20) can be expressed as:
\begin{equation}
\begin{split}
\left ( b_{i,t}-c_{i,t}\right )\left ( N_j-K_j\right )\left ( 1-f^{-1}\left ( b_{i,t}\right )\right )^{N_j-K_j-1}{f^{-1}}'\left ( b_{i,t}\right )\\
-\left (1-f^{-1}\left ( b_{i,t}\right ) \right )^{n-k}=0.
 \nonumber
 \end{split}
\end{equation}

Let $b_{i,t}$ be the optimal bid of client $i$, so, $c_{i,t} = f^{-1}\left (b_{i,t} \right )$. We also introduce the differential factor $\left ( 1-c_{i,t}\right )^{n-K_j-1}$.
\begin{equation}
\begin{split}
\left ( b_{i,t}-c_{i,t}\right )\left ( N_j-K_j\right )\left ( 1-c_{i,t}\right )^{N_j-K_j-1}{f^{-1}}'\left ( b_{i,t}\right )\\
-\left (1-c_{i,t} \right )^{N_j-K_j}=0.
\end{split}
\end{equation}
Equation (22) is a full differential equation, and the solution to the full differential equation is as follows:
$$
\left (1-c_{i,t} \right )^{N_j-K_j-1}db_i-\left ( b_{i,t}-c_{i,t}\right )\left ( N_j-k_j\right )dc_{i,t}=0
$$
\begin{multline*}
\left (1-c_{i,t} \right )^{N_j-K_j}+b_id\left ( 1-c_{i,t})\right )^{N_j-K_j}+\\
\left ( 1-c_{i,t}\right )^{N_j-K_j-1}c_{i,t}\left ( N_j-K_j\right )=0
\end{multline*}
\begin{multline*}
d\left ( \left ( 1-c_{i,t}\right )^{N_j-K_j}\times b_{i,t}\right )+\\ \frac{N_j-K_j}{N_j-K_j+1}d(1-c_{i,t})^{N_j-K_j+1}
-d(1-c_{i,t})^{N_j-K_j}=0
\end{multline*}
\begin{multline*}
\left ( \left ( 1-c_{i,t}\right )^{N_j-K_j}\times b_{i,t}\right )+\\\frac{N_j-K_j}{N_j-K_j+1}(1-c_{i,t})^{N_j-K_j+1}
-(1-c_{i,t})^{N_j-K_j}=0
\end{multline*}
$$
b_{i,t}=\frac{1}{N_j-K_j+1}+\frac{N_j-K_j}{N_j-K_j+1}\times c_{i,t}+C
$$
where $C$ is a constant, set $C$ to $0$ , so the optimal bid is:
\begin{equation}
b_{i,t}=\frac{1}{N_j-K_j+1}+\frac{N_j-K_j}{N_j-K_j+1}\times c_{i,t}.
\end{equation}

\subsection{Auction-based clients selection algorithm}
In this subsection, we describe our proposed edge clients selection algorithm in detail. From the server's perspective, the edge clients selection algorithm's function is to select appropriate edge clients to participate in federated learning training, thereby accelerating the global model's convergence rate. From the perspective of edge clients, the client selection algorithm's goal is to maximize the benefits for users and avoid excessive energy consumption. So, we designed an auction-based edge clients selection algorithm, as shown in Algorithm 1.  Algorithm 1 has three main stages, gradient-based clustering, auction-based selection, and federated training.

The server initializes the model parameters and the sample threshold $s_{mm}$. All clients randomly select $s_{mm}$ local samples to calculate the gradient-based on $w$, repeat $T_0$ times, and send the gradient average to the server. The purpose of this is to obtain local data distribution. Based on local data distribution, the clients are divided into $J$ groups. So far, the first stage is completed.

Our auction model consists of two steps: in the first step, each client calculates its own cost according to formula (6-11) and then bids according to formula (20). Then, the server randomly selects a group $js$, and selects the $k_j$ clients with the lowest bid as the winner in the group $js$. Then, let the smallest local data size among the winners as the threshold $s_{min}$. In the second step, each group is an auction system, clients with local data size greater than $s_{min}$ get the right to participate in the auction. And then, the server determines the winners in each group according to the client's bid. So far, the second stage is completed.

In the final stage, federated training is carried out. The winning clients train on local data and send the local update model weights to the server. Finally, the server aggregates the client model update and then repeats the second and third stages until the model converges or reaches the specified communication rounds.
\begin{algorithm}[]
\caption{Auction Federated Learning Based on Clustering Strategy}
\LinesNumbered
\KwIn{list of all clients $NL$, number of clients $N$, number of clusters $L$, and  proportion of selected clients $Ratio$}
\KwOut{list of selected clients $DL$}
Server broadcasts $w$, $s_{mm}$ to all clients \;
\For{each client $k \in NL$}{
    \For{$t_0=0$ to $T_0$}{
        Client $k$ selects $s_{mm}$ samples from local data\;
        Client $k$ computes its gradient $\nabla f(w, \xi_{t_0}^{k})$\;
    }
    Client $k$ computes the mean value of $\nabla f(w, \xi_{t_0}^{k})$\;
    Client $k$ sends its mean value of gradient to server\;
}
Server clusters clients into $J$ groups according to client's gradient\;
Server computes the number of selected clients, $K$\;
\For {$t=0$ to $T$}{
Server broadcasts training request\;
    \For {each client $k \in NL$}{
        Client $k$ computes its cost according to (9-14)\;
        Client $k$ computes bid $b_{k,t}$ according to (23)\;
        Client $k$ sends its bid $b_{k,t}$ back to server\;
    }
    Server computes the number of selected clients in each group $j$: $K_j = K/J$\;
    Server randomly selects a group $js$\;
    Server selects $K_j$ clients with lowest bid in group $js$: $KL_js$\;
    Server computes minimum local data size of $KL_{js}$, $s_{min}$\;
    \For{$j=0$ to $J$}{
        \For{each client $k \in$ group $j$}{
            \If{$k_s\geq s_{min}$}{
                $JL_{j,t} =JL_{j,t}\bigcup k$
        }
        }
        Server selects $K_j$ clients with lowest bid in $JL_{j,t}$: $KL_{j,t}$\;
        $DL_t = DL_t\bigcup KL_j$\;
    }
    \For{each client $k \in DL$}{
        % Client $k$ selects $s_{min}$ samples from local data\;
        Client $k$ trains on local data\;
        Client $k$ updates model weights, $w^{k}_{t+1}$\;
        Client $k$ sends $w^{k}_{t+1}$ back to server\;
    }
    Server aggregates model parameters $w_{t+1}$ (5)\
}
\end{algorithm}

\section{EVALUATION}
\subsection{Simulation set up}
We have implemented our scheme in a federated learning simulator developed from scratch using PyTorch under a device with a 3.0GHz CPU frequency. We choose three classic picture datasets for dataset selection: MNIST, Fashion MNIST (represented by FMNIST), and CIFAR-10. These datasets all contain ten types of data samples, and the former two datasets both have 70,000 images, 60,000 for training, and 10,000 for testing. The latter has 60,000 images, 50,000 for training, and 10,000 for testing. In terms of the AI model, we trained three different models, including CNN for MNIST\footnote{The CNN for MNIST has 10 layers with the following structure: 5×5×10 Convolutional
→ 2×2 MaxPool → 5×5×20 Convolutional → Dropout → 2×2 MaxPool → Flatten
→ 320×5 Fully connected → dropout → 50×10 Fully connected → softmax.}, CNN for Fashion MNIST\footnote{The CNN for Fashion MNIST has 9 layers with the following structure: 5×5×16 Convolutional
→ Batch Normalization → 2×2 MaxPool → 5×5×32 Convolutional → Batch Normalization → 2×2 MaxPool → Flatten → 1568×10 Fully connected → softmax.
} and CNN for CIFAR-10\footnote{The CNN for CIFAR-10 has 8 layers with the following structure: 5×5×6 Convolutional
→ 2×2 MaxPool → 5×5×16 Convolutional → Flatten → 400×120 Fully connected → 120×84 Fully connected → 84×10 Fully connected → softmax.
}. Besides, our scheme mainly considers clients' selection, using FedAvg and FedProx respectively when modeling aggregation. Therefore, we mainly compare with the randomly selected FedAvg and FedProx, which are represented by Random$\_$FedAvg and Random$\_$FedProx, respectively.

\begin{table}[]
    \caption{Simulation parameters}
    \centering
    \begin{tabular}{p{7cm}p{1cm}}
        \hline
        Parameter & Value\\
        \hline
        Sample windows size, $s_{mm}$  & 50                      \\
        Energy consumption per 100 samples, $\varrho$  & 0.2                      \\
        Cost parameter related to the residual energy, $\phi$  & 0.5                     \\
        Cost parameter related to local samples, $\vartheta$  & 0.5              \\
        Weight parameter related to local samples, $\chi$  & 0.7          \\
        Cost parameter related to communication rounds, $a$  & 2                    \\
        Weight parameter related to communication rounds, $\varsigma$  & 0.3                      \\
        Weight parameter related to the resource cost, $ \alpha $  & 0.7                     \\
        Weight parameter related to the service cost, $\lambda $  & 0.3                     \\
        \hline
    \end{tabular}
    \label{cnn for mnist}
\end{table}

\textbf{Data Distribution at Different clients:} For the Non-IID setting, we use the \cite{Wang} setting method, that is, the percent $\nu$ of the samples stored by each client is the same label, and the remaining data samples are randomly sampled. In our simulation, $\nu$ is set to: 1, 0.8, 0.5. Besides, for imbalance setting, the number of local samples for each client is between $\varpi/6$ and $2\varpi$, Where $\varpi$ is the average value that samples of each client can allocate. Taking the 100 clients scenario and MNIST dataset as an example, the value of $\varpi$ is 600 at this time, and each client has at least 100 data samples and at most 1200 data samples. Then, each client's local data, of which 80\% is used for training, 10\% is used for verification, and the last 10\% is used for testing.

\textbf{Energy at different mobile clients:} We assume that in the system, each client's battery capacity is the same so that the percentage of energy represents the remaining energy of each client. For each client's initial energy, we considered two scenarios: case1: Each client has the same energy size; we set it to 100\%. Case2: we set the energy of all clients in the system to satisfy a normal distribution with an upper bound of 100\%, a lower bound of 50\%, a mean of 75\%, and a standard deviation of 10. In other words, the energy of all clients in the system is between 50\% and 100\%.

\subsection{Simulation results}

\subsubsection{Price and reward}

We propose two reward models in section VI. In our simulation, we choose the second reward model. Firstly, the AI model trained by the federated learning system has certain economic returns, which belong to the whole federated learning system. In other words, the server and clients share the economic returns of the AI model. Secondly, we divide the economic returns equally according to the communication rounds required by the training model. Thirdly, we mainly measure all clients' average bids for clients' prices in each round of the system. The reward mainly includes two parts: server-side and clients side. For the server-side, we record the server's reward with the number of communication rounds. For the client's side, we mainly consider the sum of each round of training's benefits.

\begin{figure}[htbp]
\begin{minipage}[t]{0.235\textwidth}
\includegraphics[width=4.3cm]{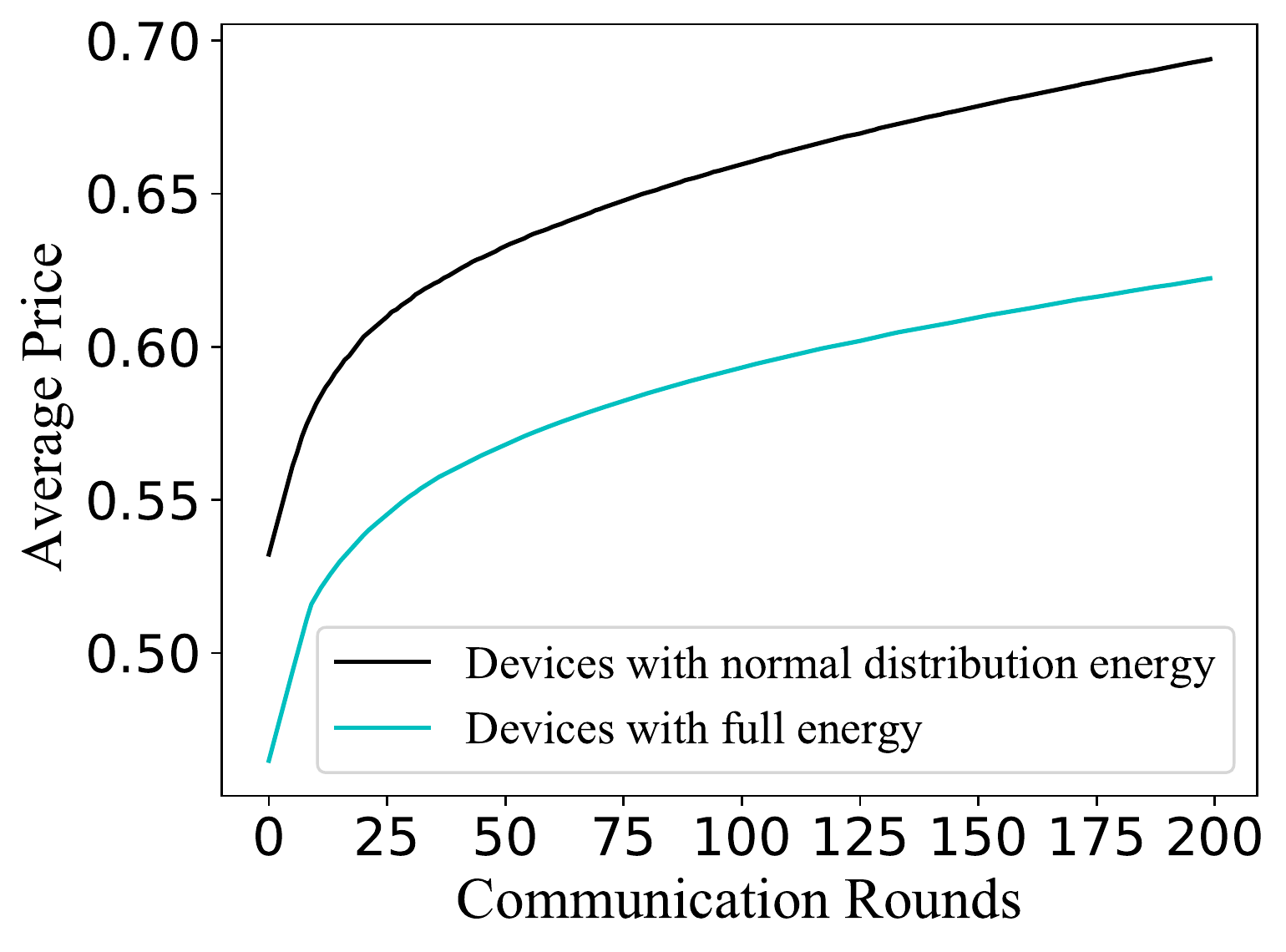}
\end{minipage}%
\begin{minipage}[t]{0.235\textwidth}
\includegraphics[width=4.3cm]{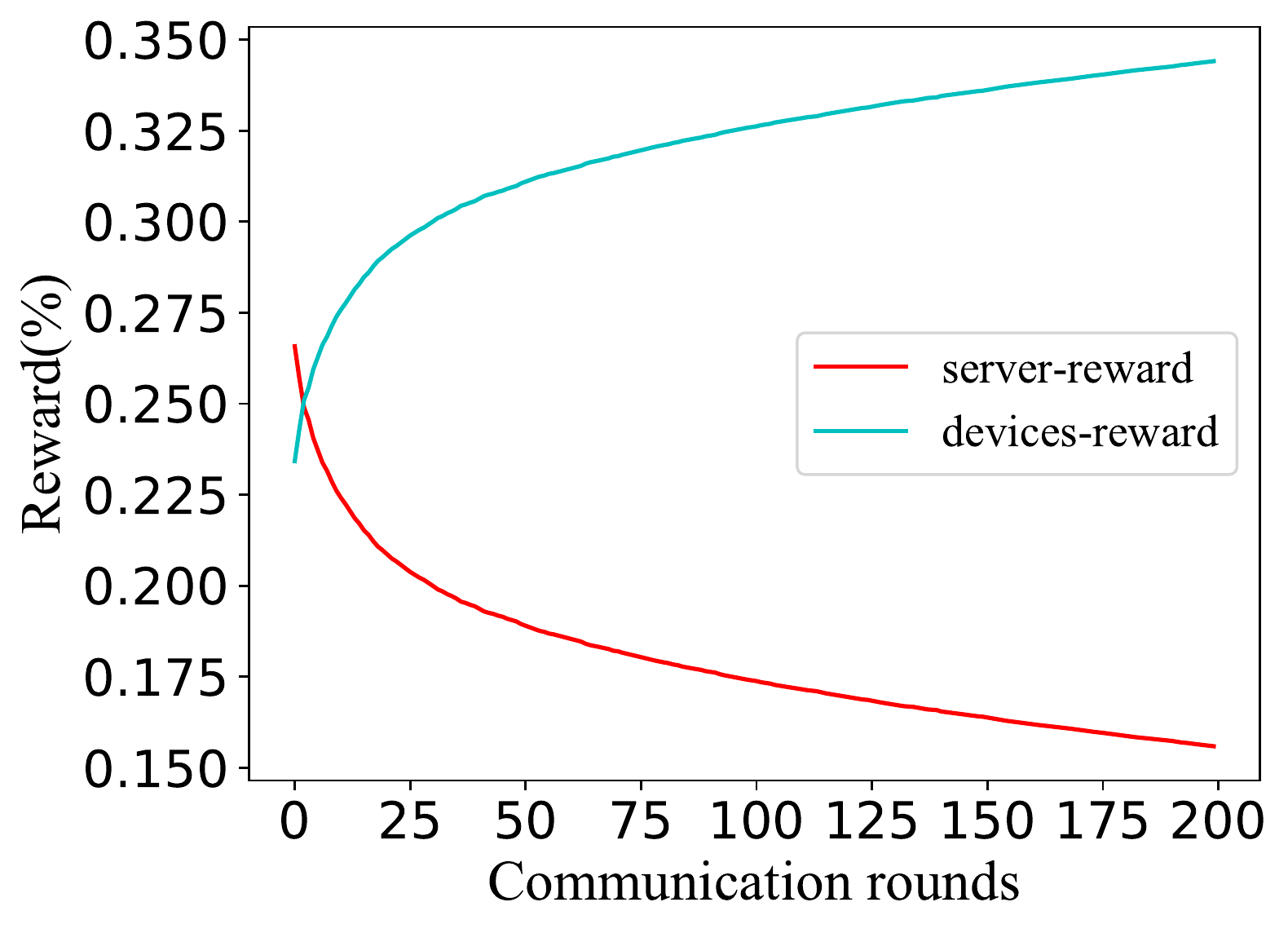}
\end{minipage}
\caption{Price and reward vs communication rounds on MNIST data. (price(left), reward(right)).}
\end{figure}

Our simulation results show that with the increase of the number of communication rounds, clients' bidding is also increasing shown in Fig. 8. As the number of training rounds increases, the remaining energy of the client is decreasing, so the cost and bid will also increase. Besides, it also shows that the client's bid will be lower when the battery is full, which is also in line with economic theory. Fig. 8 also shows that with the increase of communication rounds, the server's reward per round decreases. This is because the bidding of clients is increasing, and the proportion of clients in reward is also increasing, so the server-side reward is decreasing.

\subsubsection{Convergence rate}
The convergence rate of the global model is represented by the test accuracy rate vs. the number of communication rounds. This simulation scenario mainly contains 100 clients, and in each round of training, 10\% of the clients are selected to participate in the training, and then we mainly test the scheme's accuracy with the change of communication rounds on MNIST, fashion MNIST datasets, and CIFAR-10 datasets. The first two schemes are our proposed client selection schemes based on initial gradient clustering. The first scheme uses a random selection strategy (represented as Gradient-Cluster-Random) in each cluster, and the second scheme adopts an auction-based client selection scheme in each cluster (represented as Gradient-Cluster-Auction). The third scheme is the classic FedAvg \cite{Mcmahan}, which randomly selects clients in the system to participate in training (represented as
Random-FedAvg). The last scheme is the classic FedProx \cite{DBLP:journals/fgcs/MothukuriPPHDS21}, which also adopting random selection (represented as Random-FedProx).
\begin{figure}[htbp]
\begin{minipage}[t]{0.5\linewidth}
\includegraphics[width=4.3cm]{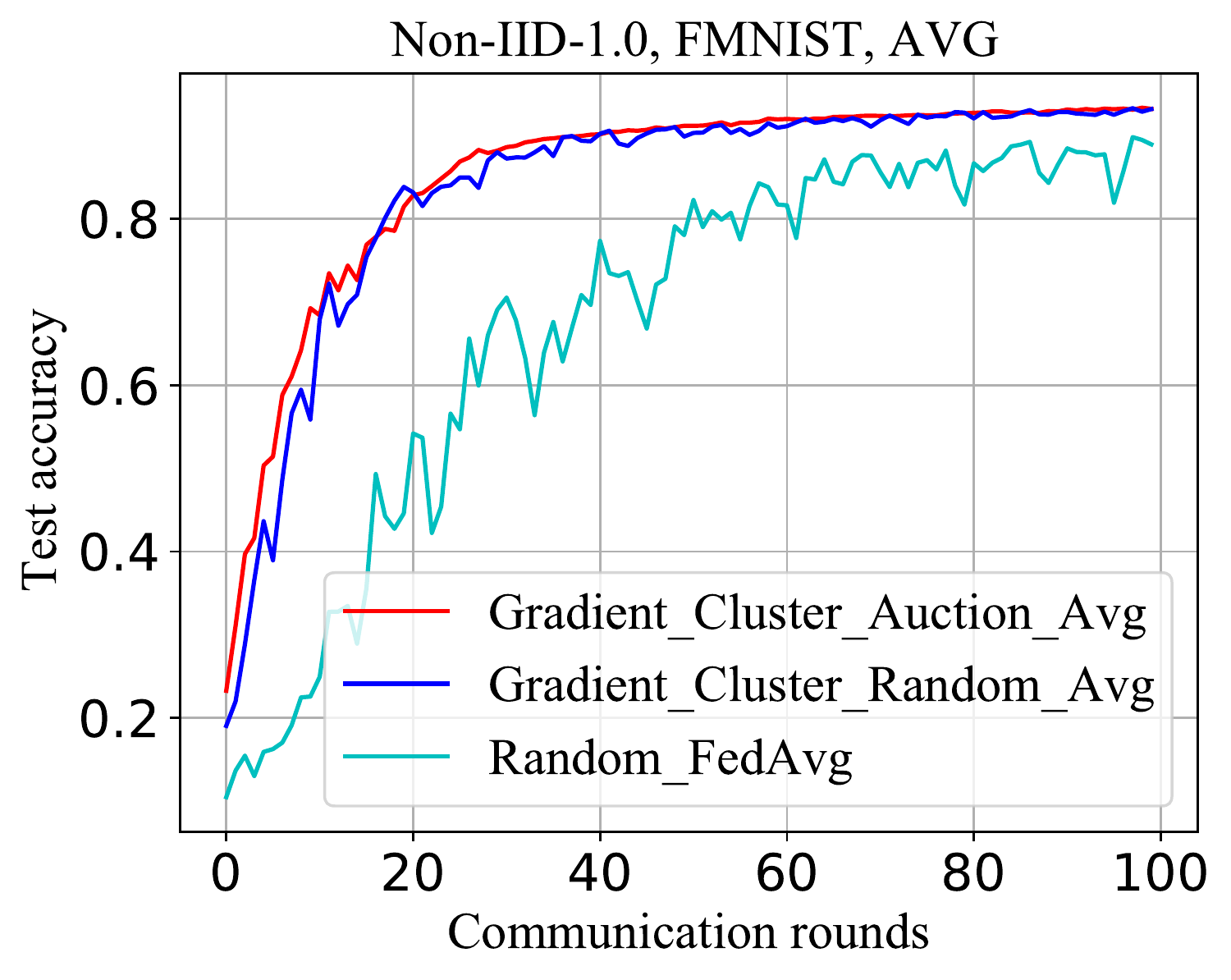}
\includegraphics[width=4.3cm]{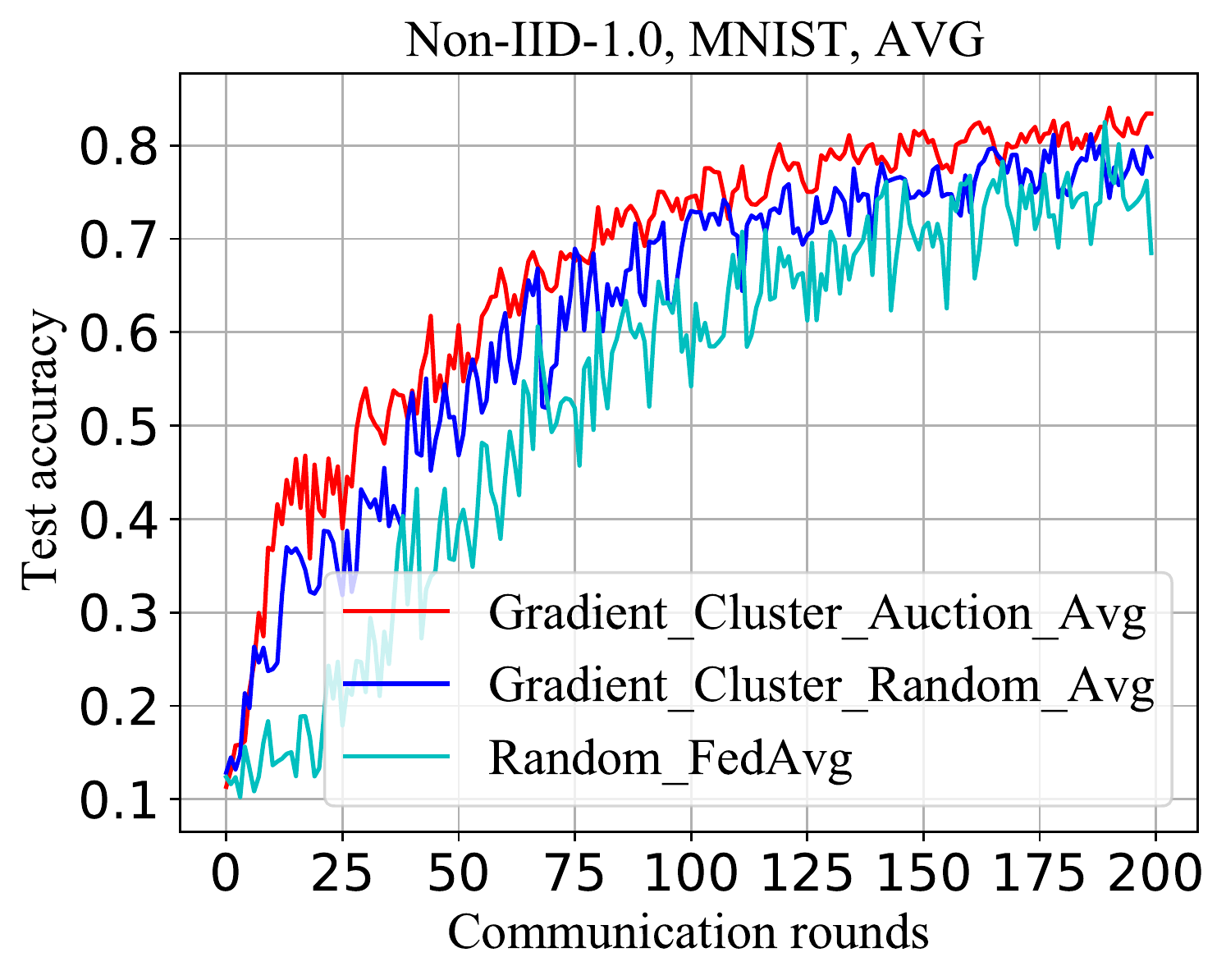}
\includegraphics[width=4.3cm]{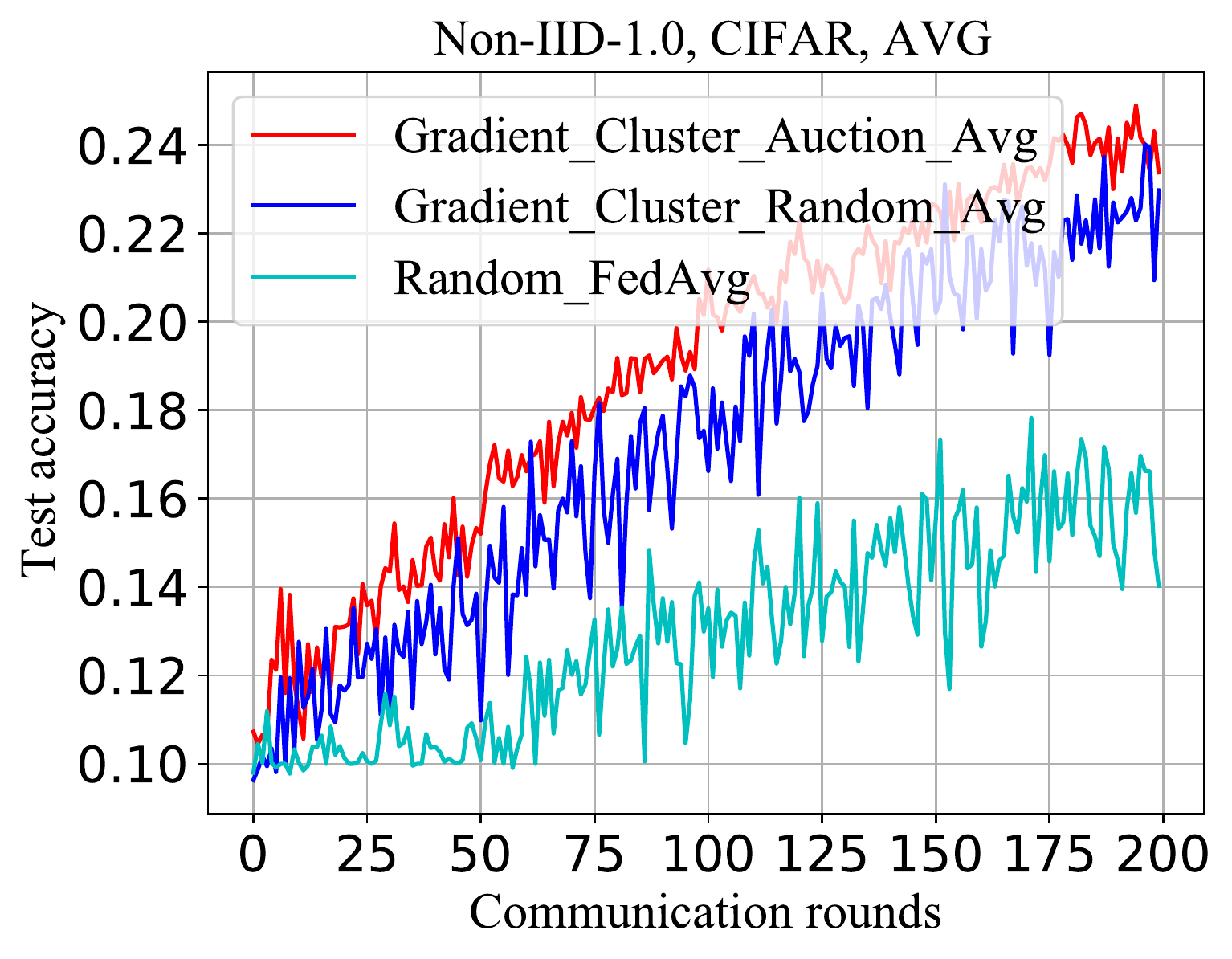}
\end{minipage}%
\begin{minipage}[t]{0.5\linewidth}
\includegraphics[width=4.3cm]{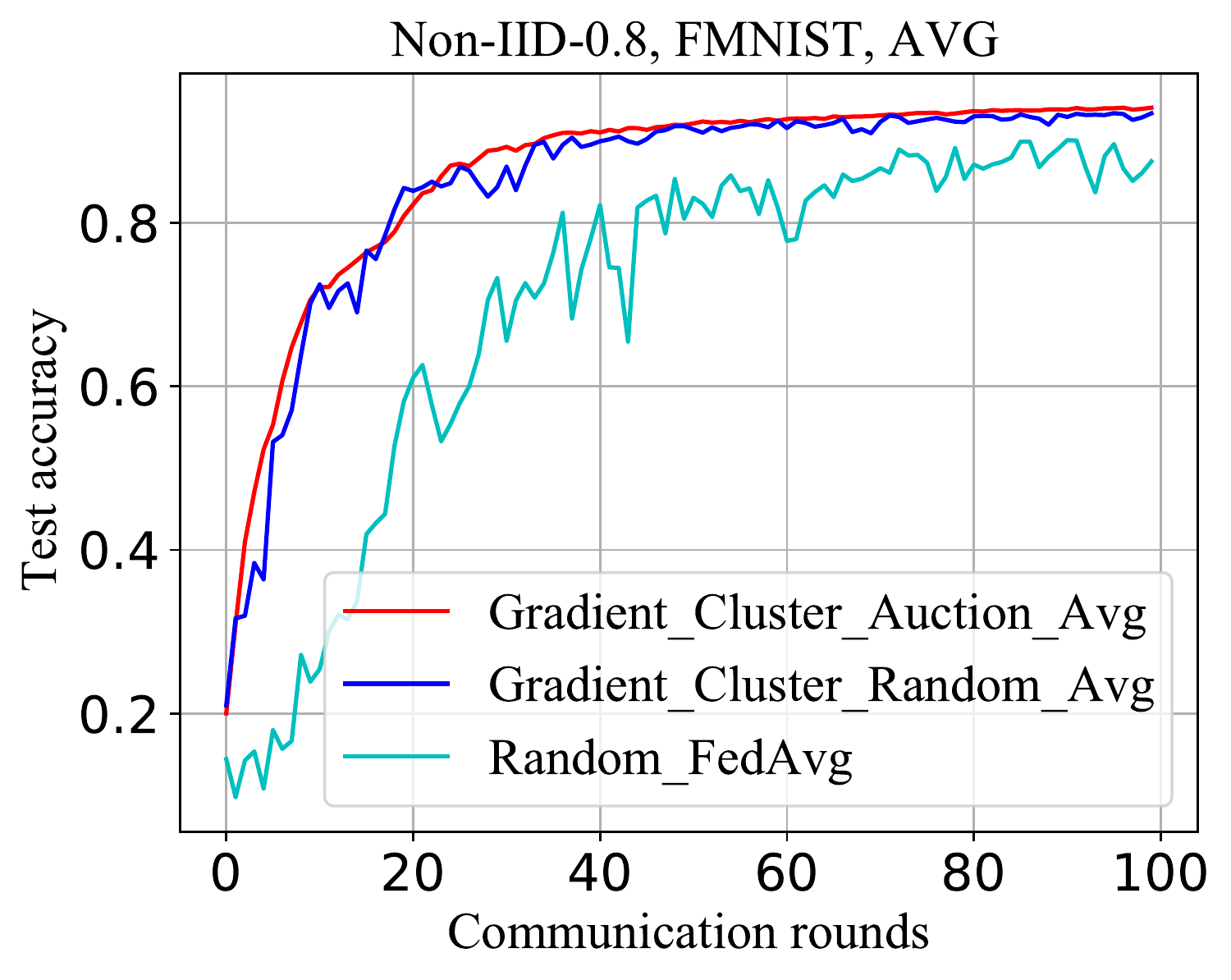}
\includegraphics[width=4.3cm]{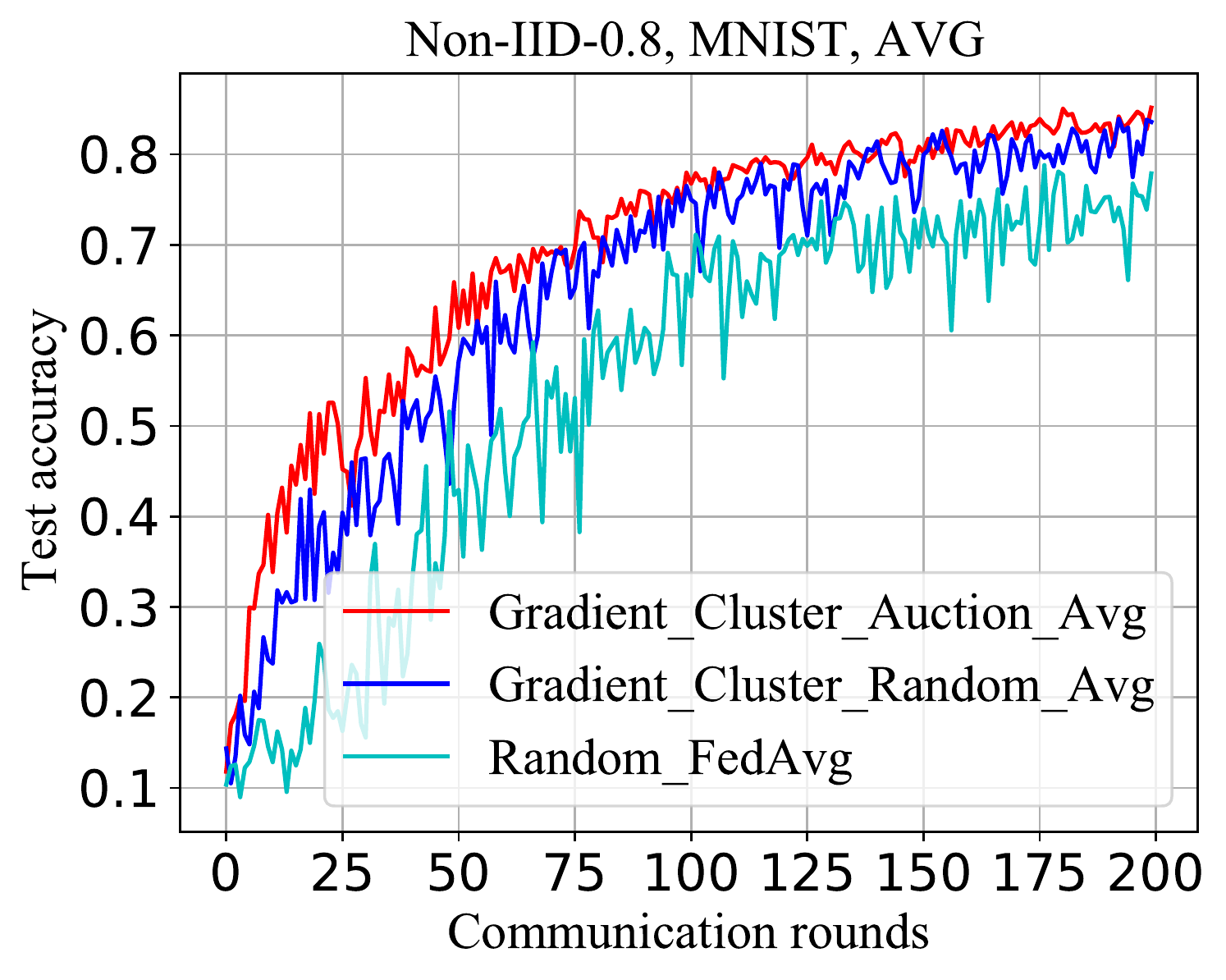}
\includegraphics[width=4.3cm]{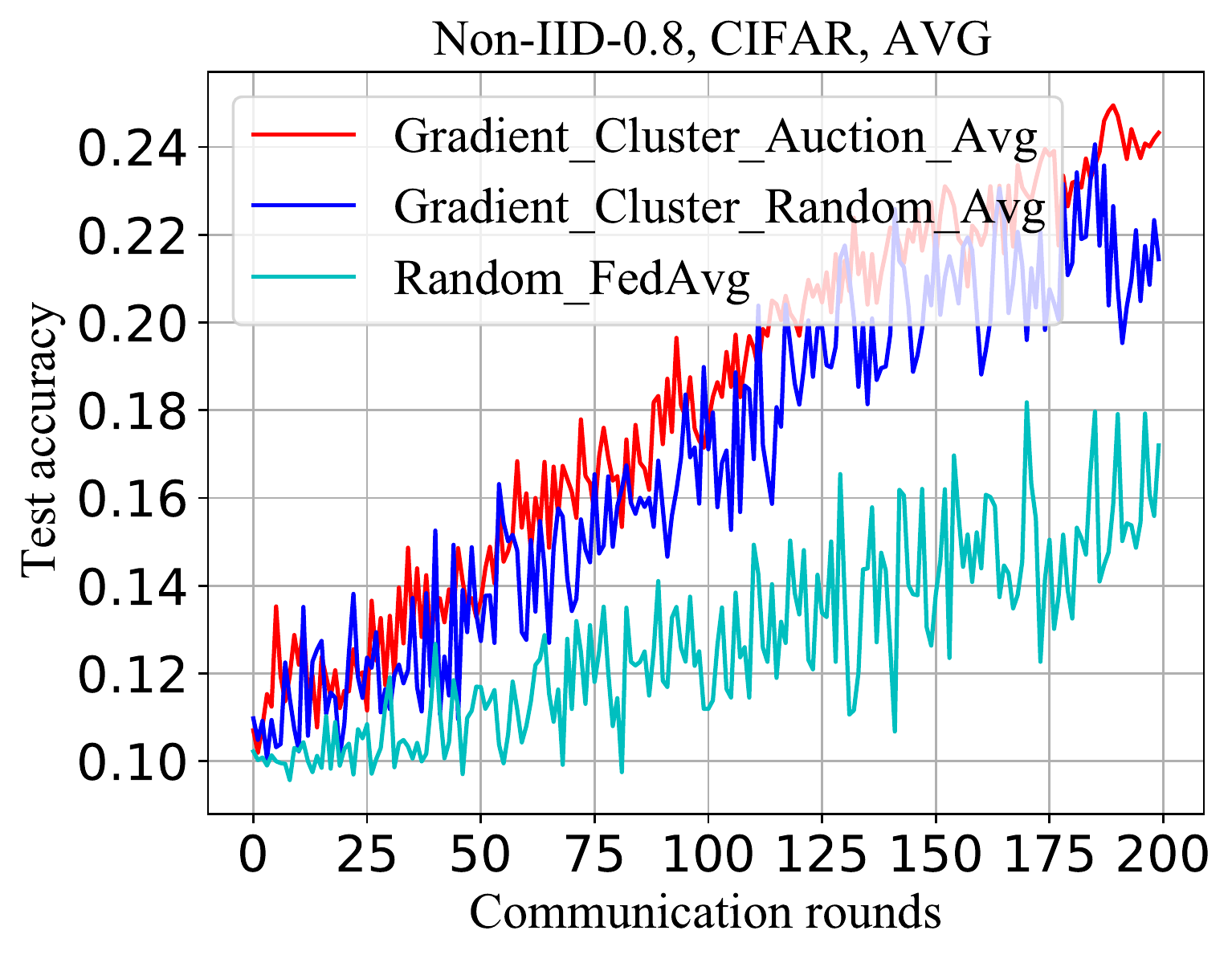}
\end{minipage}
\caption{Test accuracy v.s. communication rounds on different levels of Non-IID under 100 clients, when using AVG.}
\end{figure}

\begin{figure}[htbp]
\begin{minipage}[t]{0.5\linewidth}
\includegraphics[width=4.3cm]{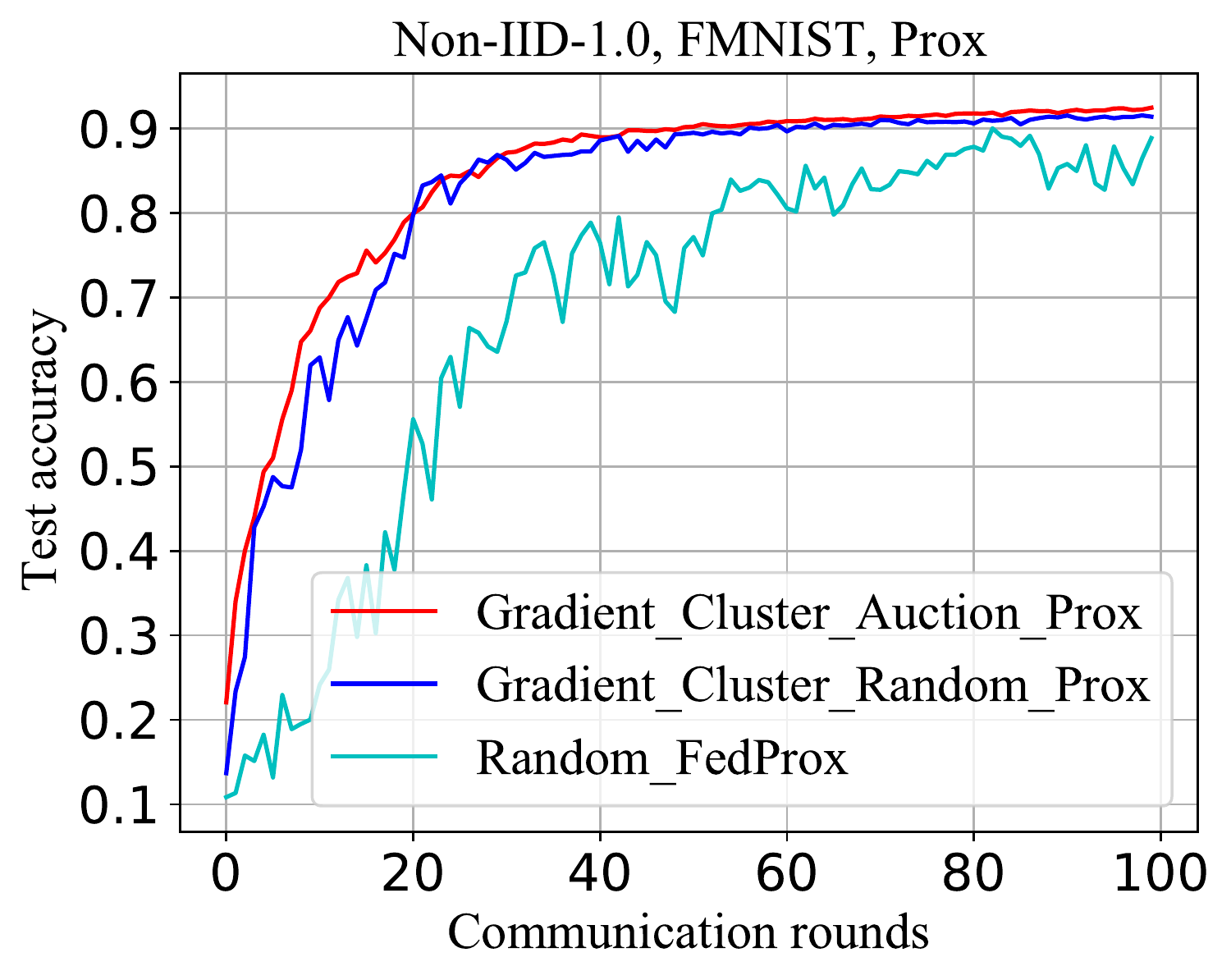}
\includegraphics[width=4.3cm]{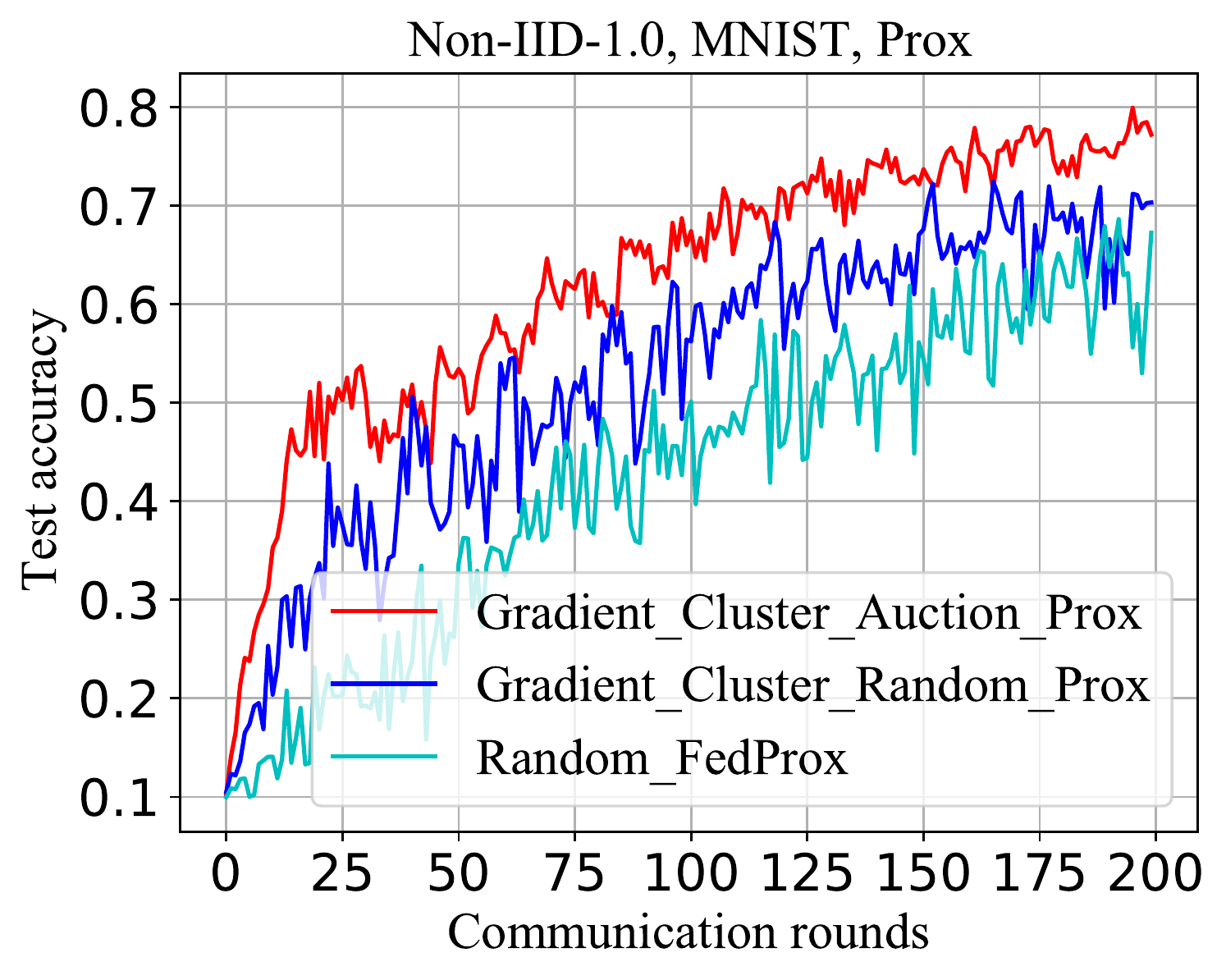}
\includegraphics[width=4.3cm]{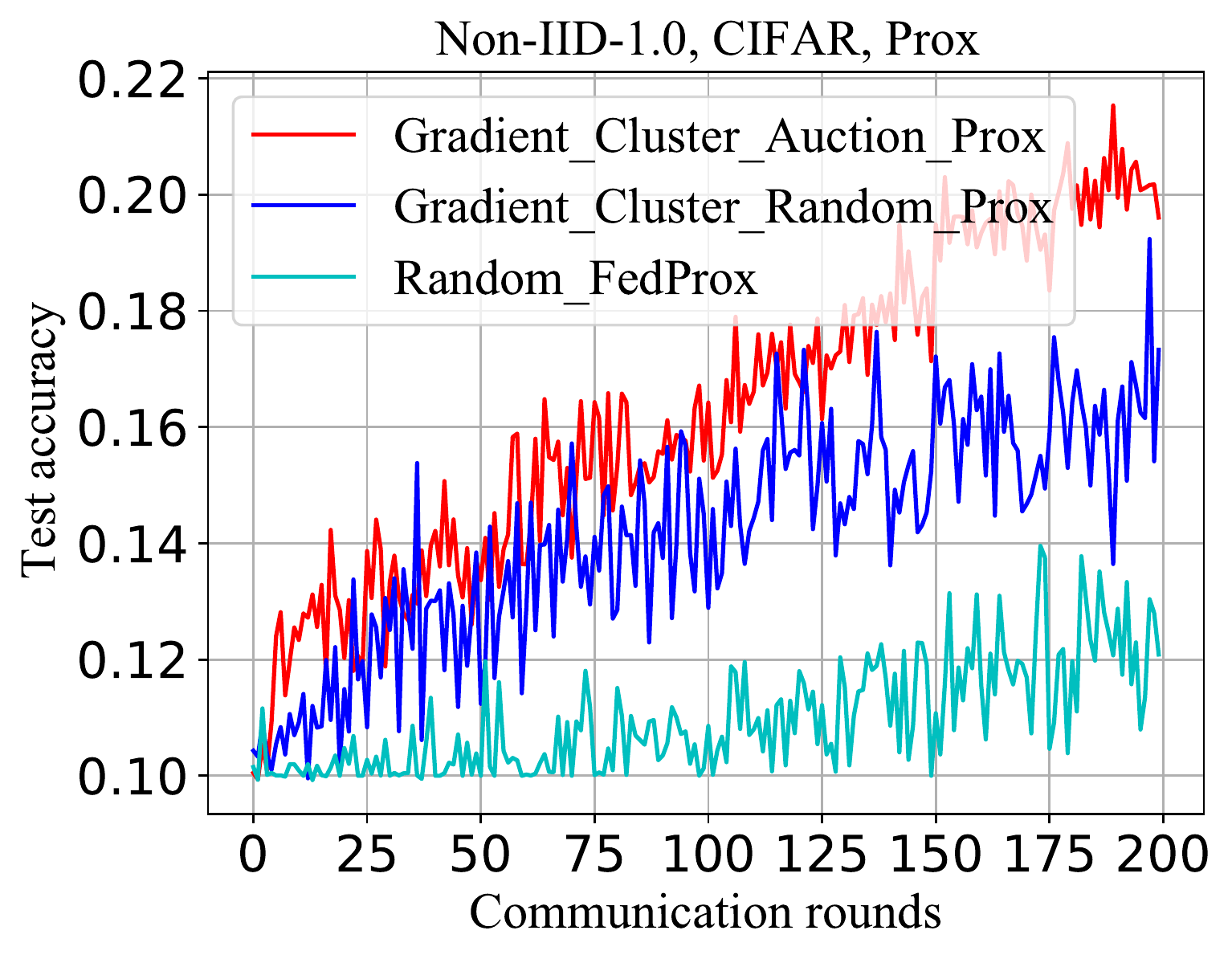}
\end{minipage}%
\begin{minipage}[t]{0.5\linewidth}
\includegraphics[width=4.3cm]{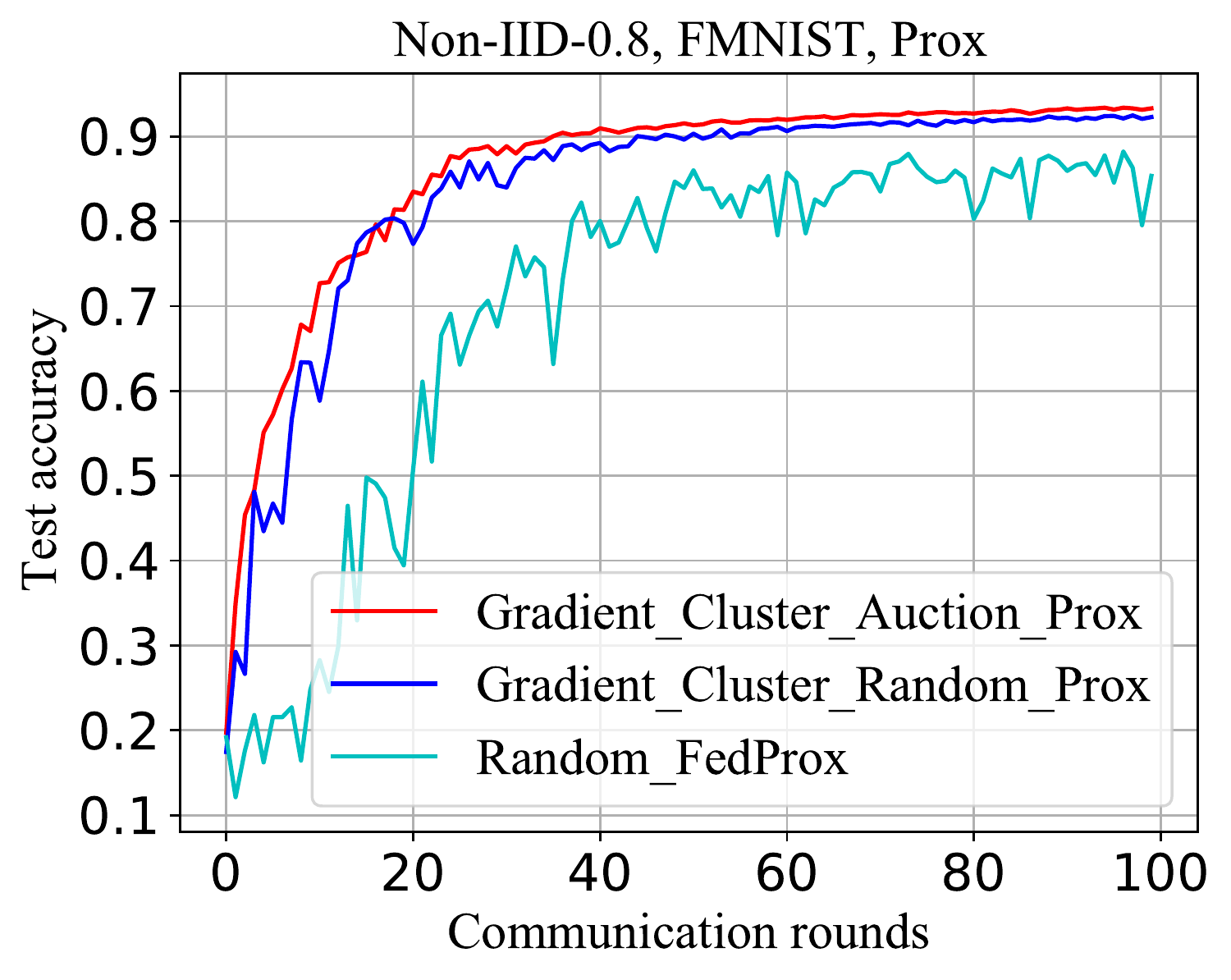}
\includegraphics[width=4.3cm]{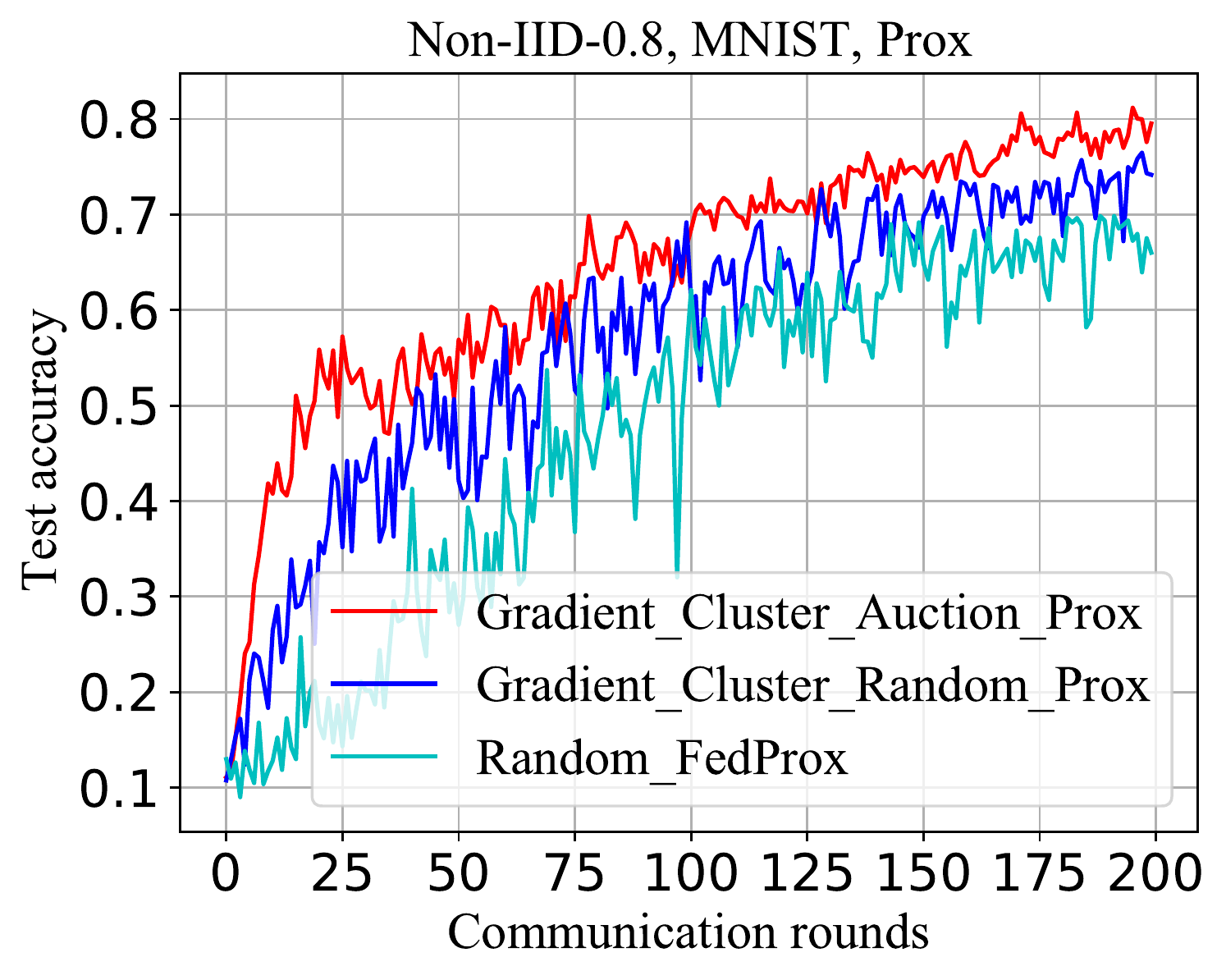}
\includegraphics[width=4.3cm]{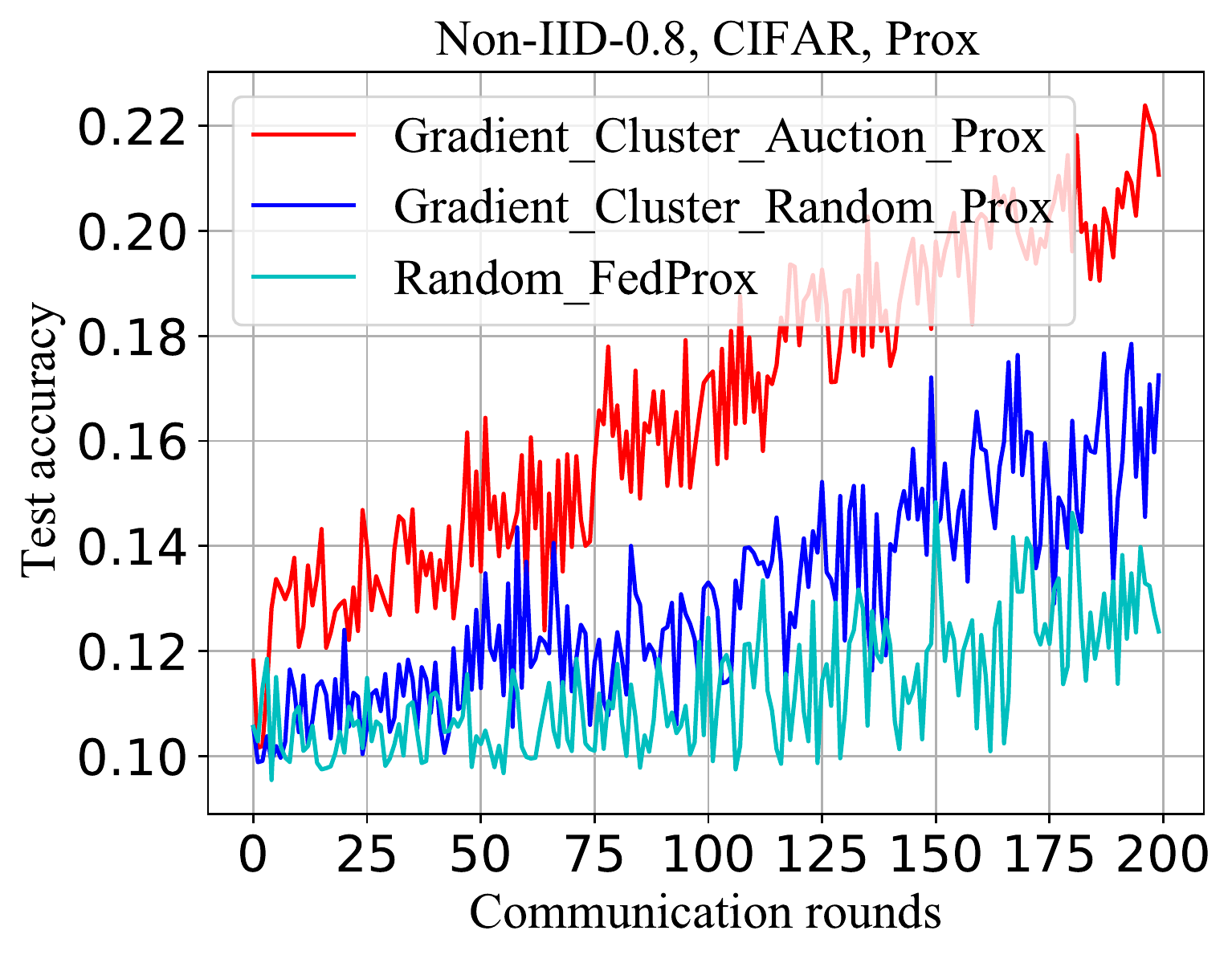}
\end{minipage}
\caption{Test accuracy v.s. communication rounds on different levels of Non-IID under 100 clients, when using Prox.}
\end{figure}

On the one hand, we test the three schemes' accuracy under two different Non-IID settings: $\nu=1$ (left) and $\nu=0.8$ (right). As shown in Fig. 5 and Fig. 6, the proposed cluster selection client scheme is better than FedAvg and FedProx in convergence rate. The reason is that the first two schemes fully consider the local data distribution of different clients when sampling clients. Through the clustering sampling strategy, a virtual data set is constructed in each round, which approximates the global distribution in distribution, thereby alleviating the influence of data heterogeneity on the global model's convergence. However, for FedAvg and Fedprox with random sampling strategy, the distribution of the selected local data set is different in different rounds, which has a negative impact on the convergence of the global model. Besides, compared to randomly selecting a client from each class after clustering, the auction-based client selection scheme can show a faster convergence rate. This is because our auction scheme fully considers the local data size when choosing clients, making the average number of training samples larger than our proposed cluster scheme. On the other hand, as shown in Fig. 7, our proposed clustering client selection scheme shows a better convergence rate when IID dominates ($\nu=0.5$). As the impact of data heterogeneity decreases, compared to the auction client selection scheme, the cluster random selection scheme can have more clients portfolio.
\begin{figure}[htbp]
\begin{minipage}[t]{0.235\textwidth}
\includegraphics[width=4.3cm]{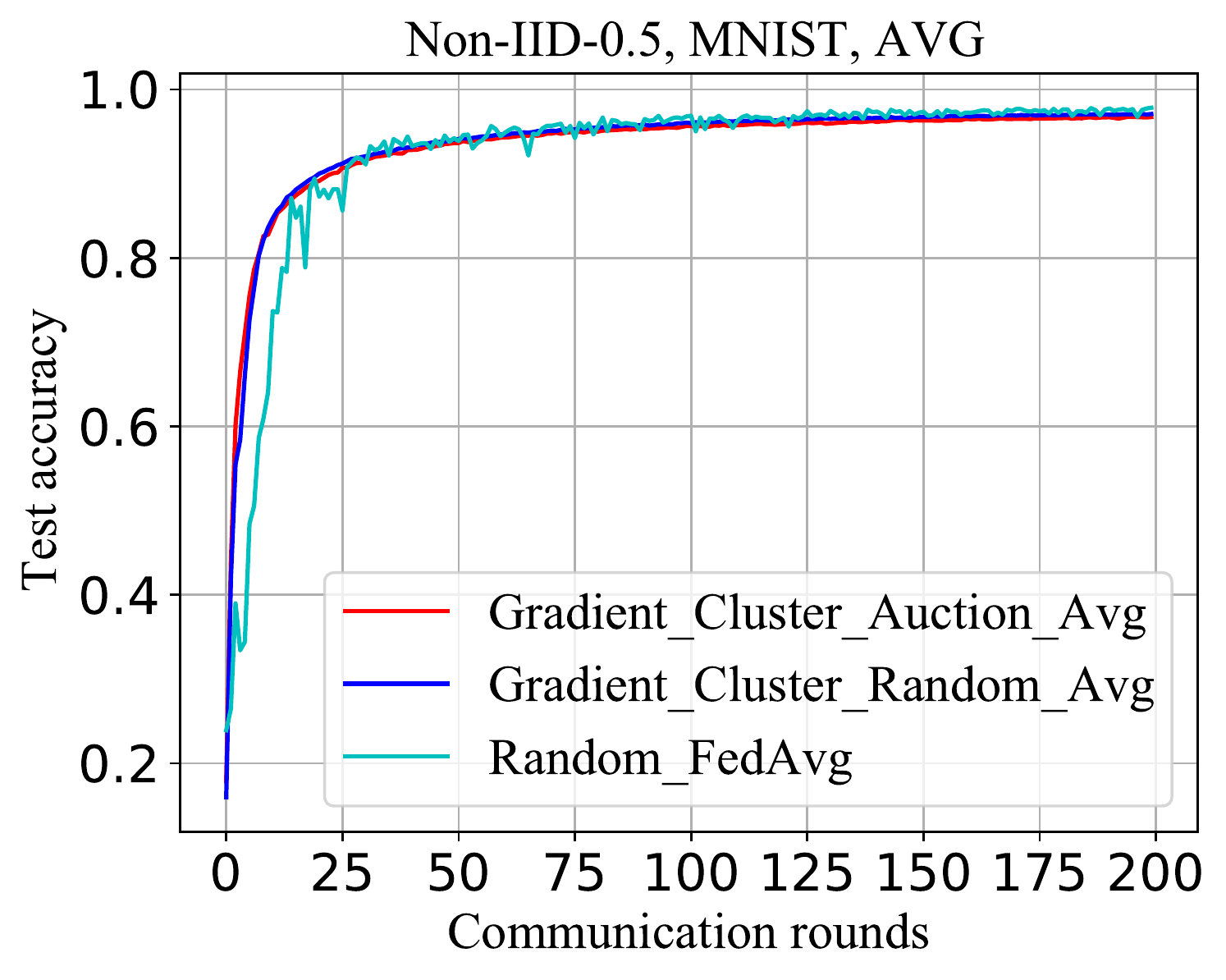}
\includegraphics[width=4.3cm]{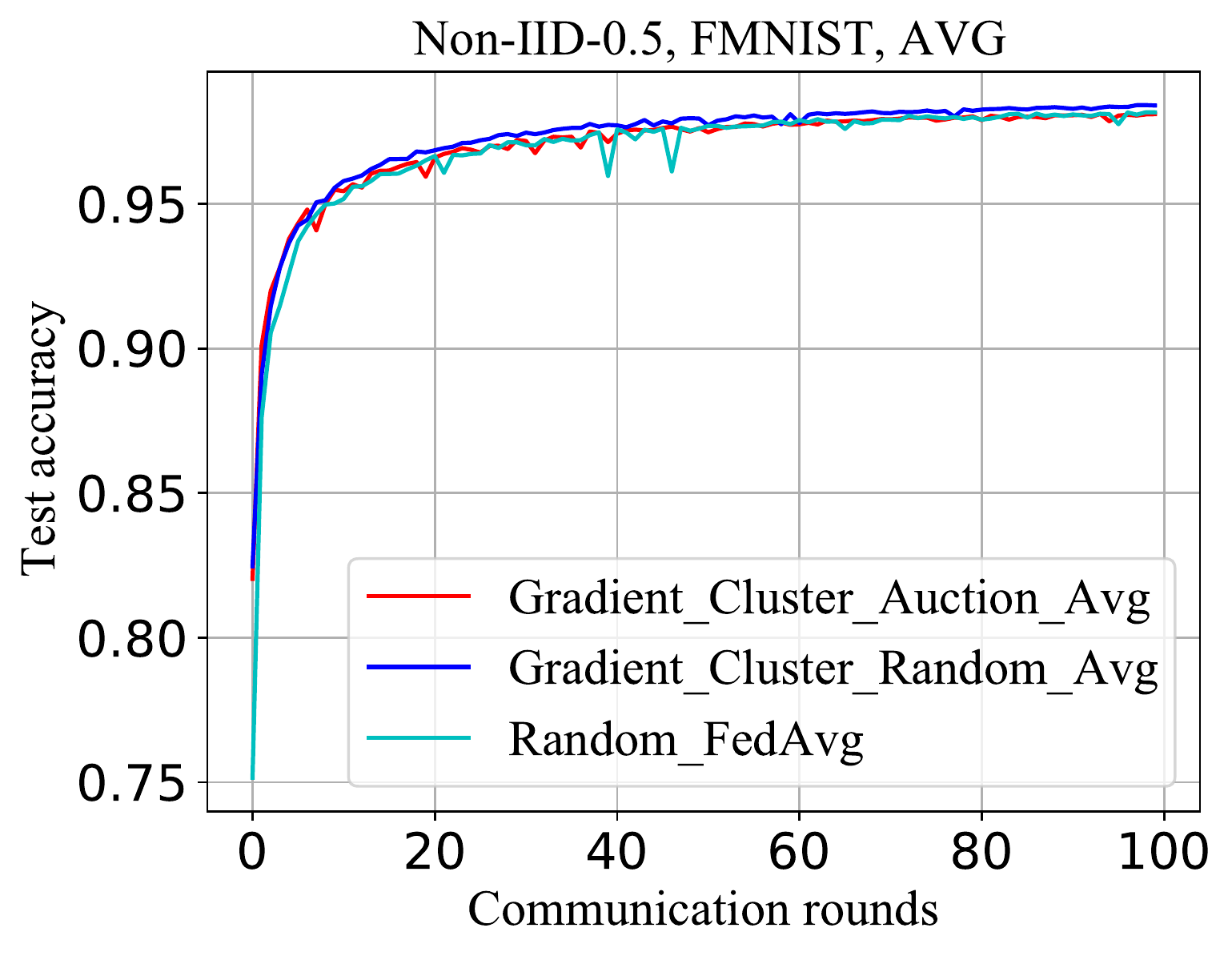}
\includegraphics[width=4.3cm]{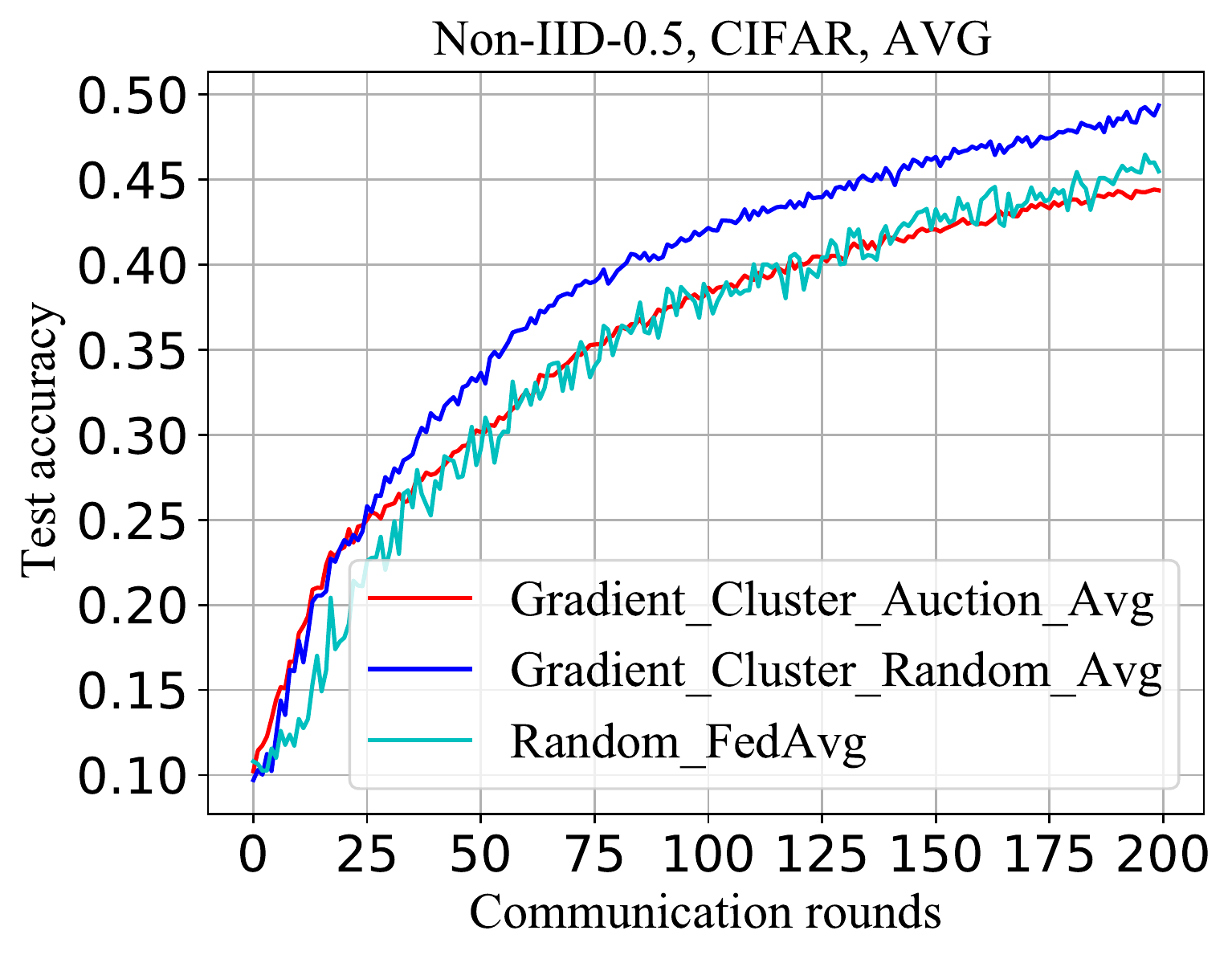}
\end{minipage}%
\begin{minipage}[t]{0.235\textwidth}
\includegraphics[width=4.3cm]{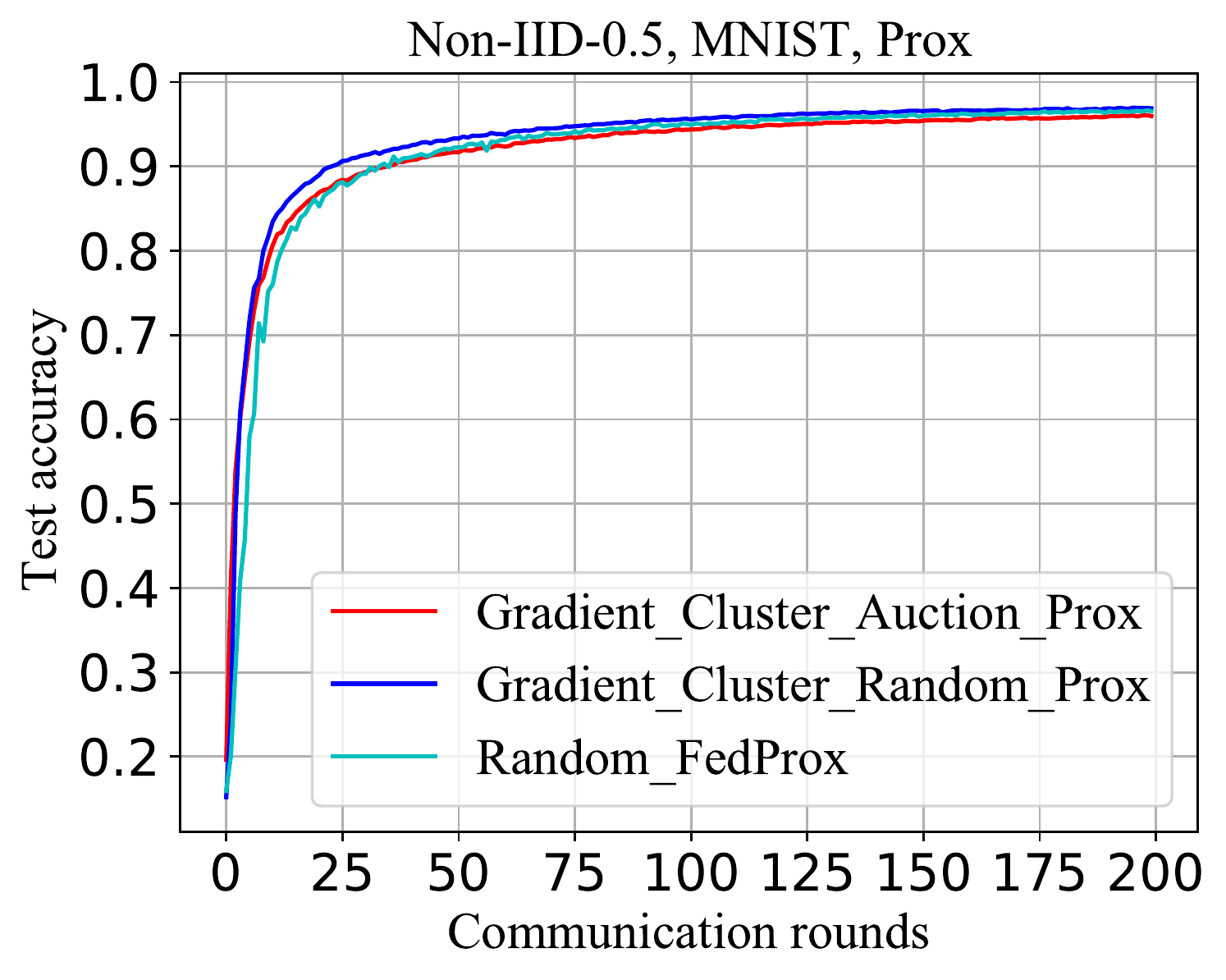}
\includegraphics[width=4.3cm]{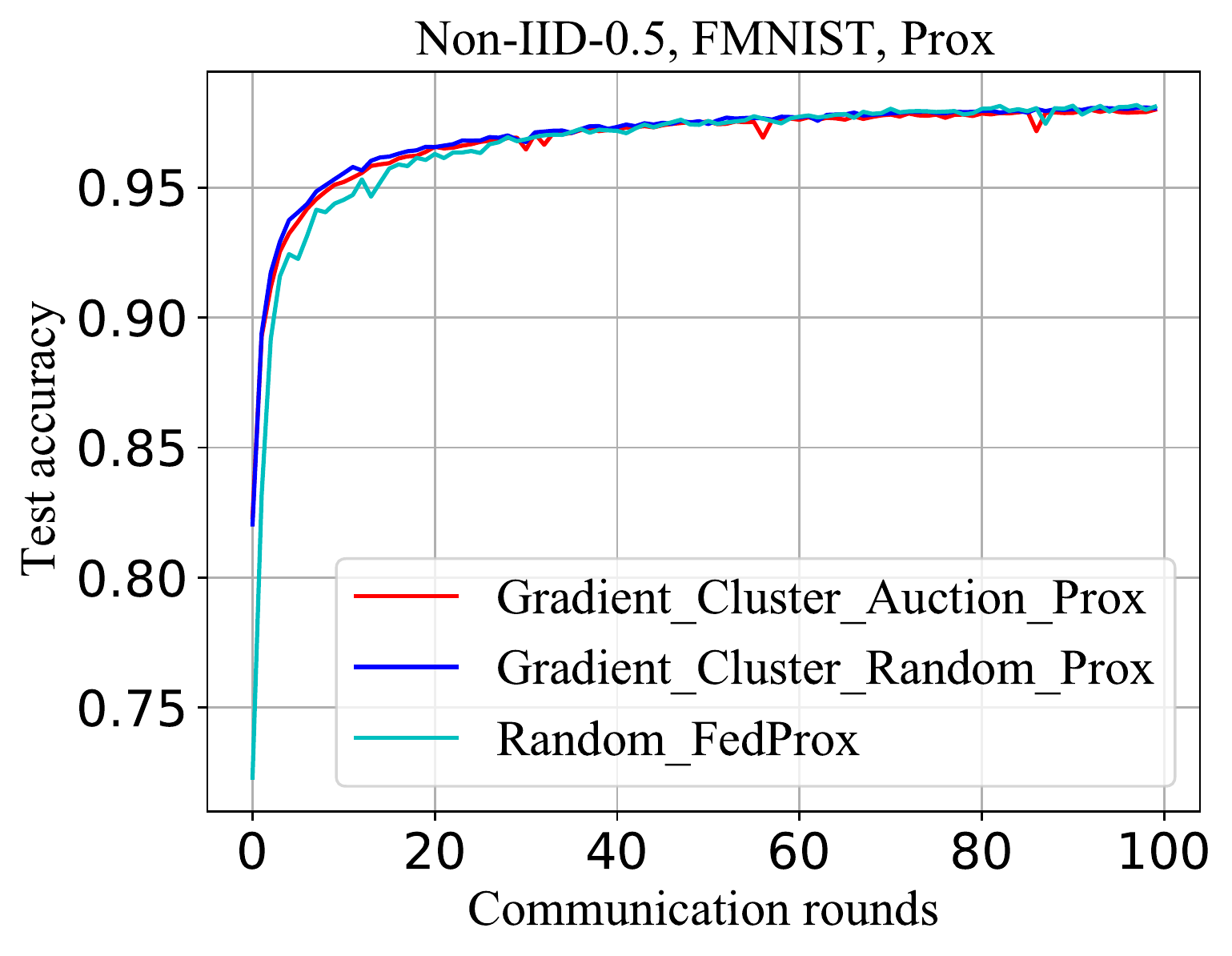}
\includegraphics[width=4.3cm]{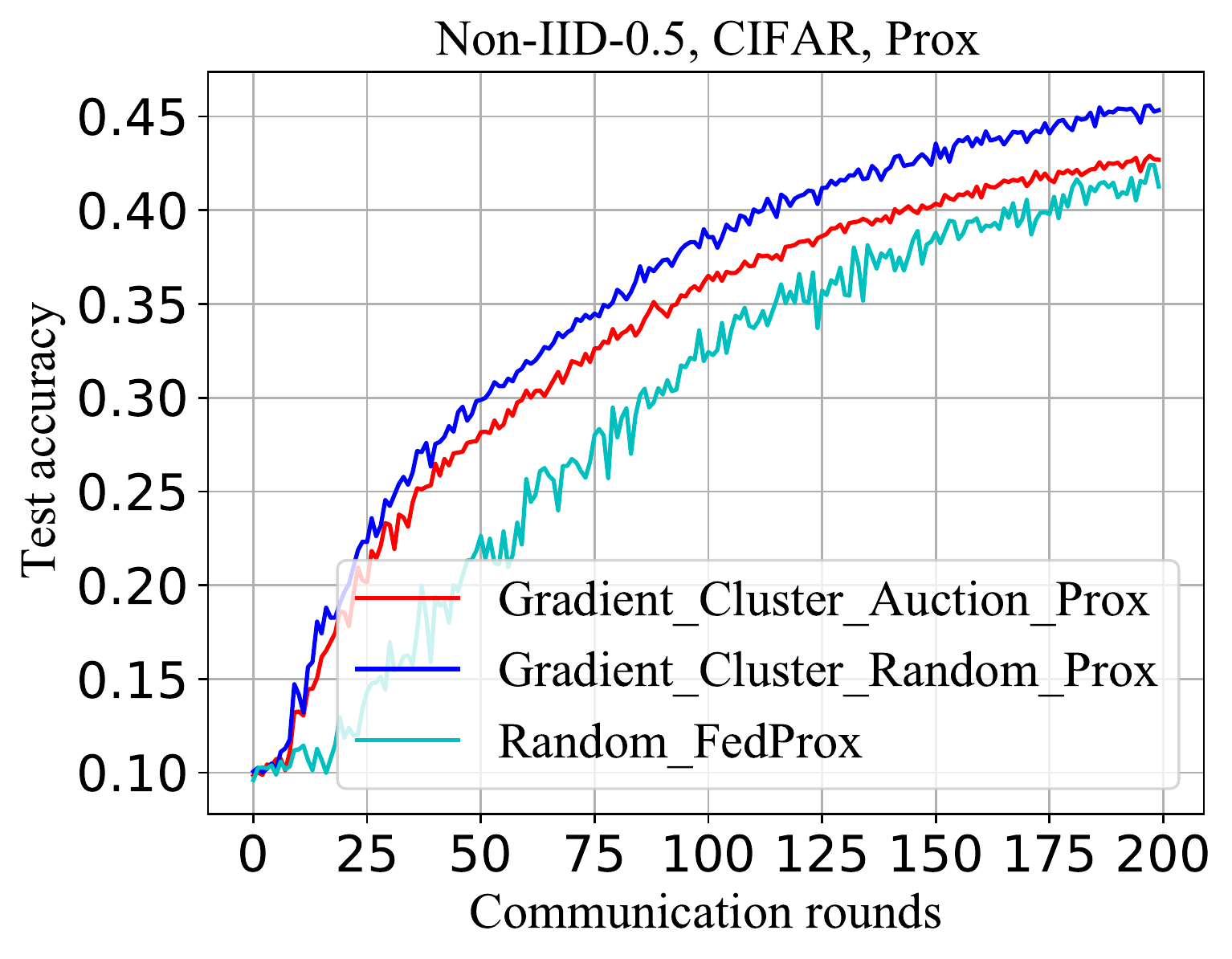}
\end{minipage}
\caption{Test accuracy v.s. communication rounds on Non-IID-0.5 under 100 clients. (Avg (left) and Prox (right)).}
\end{figure}

\begin{figure*}[!t]
\centering
\subfigure[Fashion MNIST]{\includegraphics[width=2.3in,angle=0]{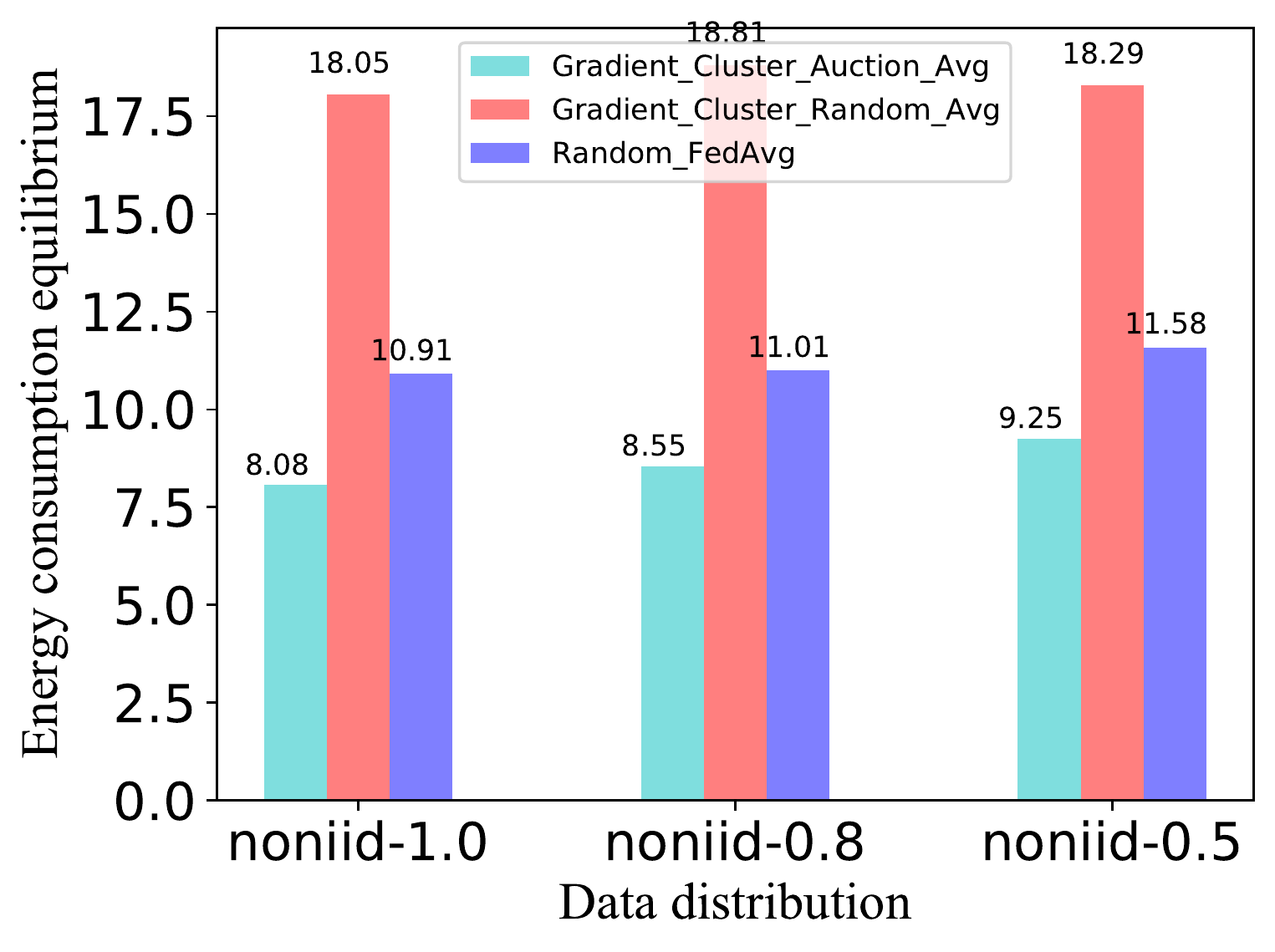}}
\subfigure[MNIST]{\includegraphics[width=2.2in,angle=0]{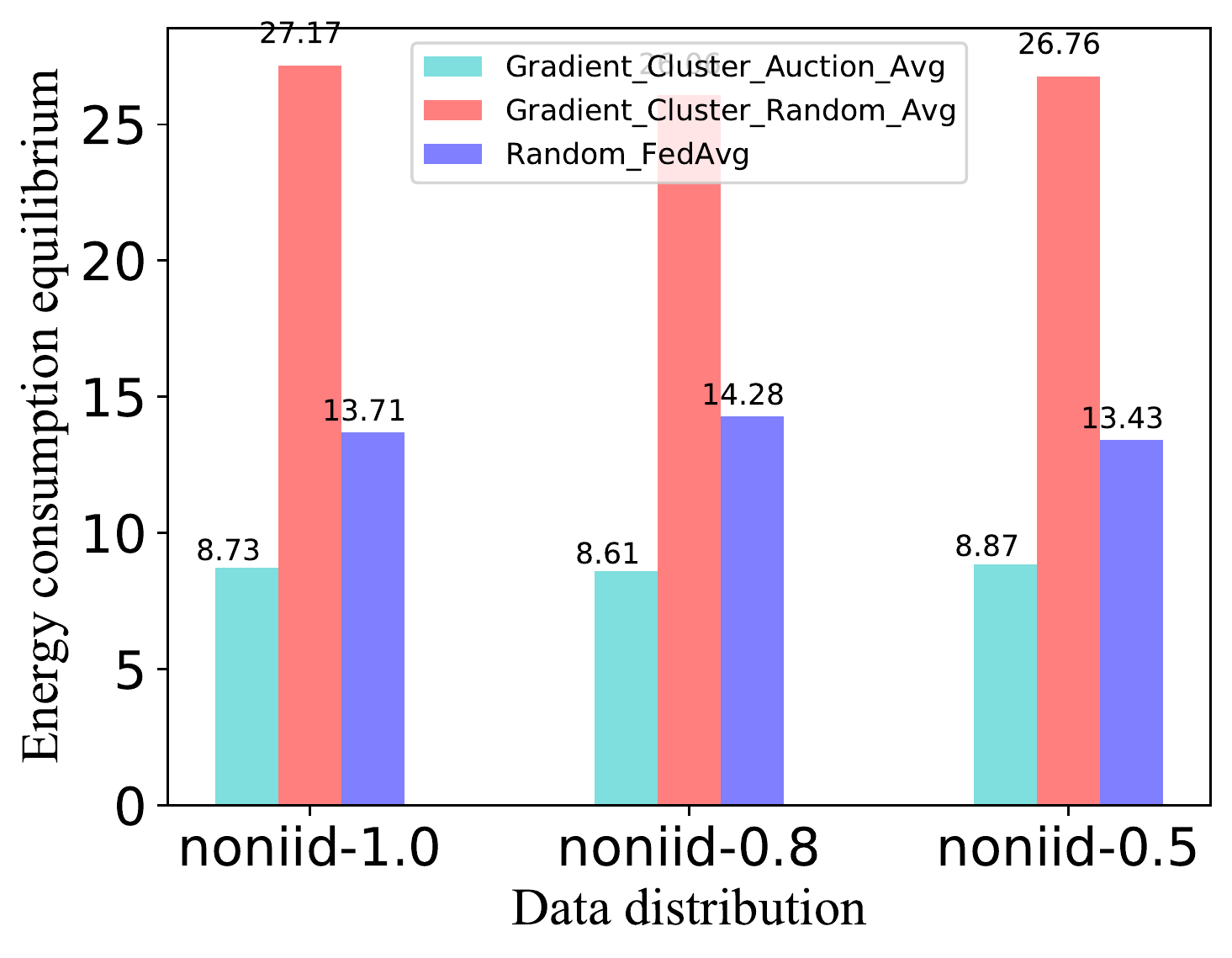}}
\subfigure[CIFAR-10]{\includegraphics[width=2.2in,angle=0]{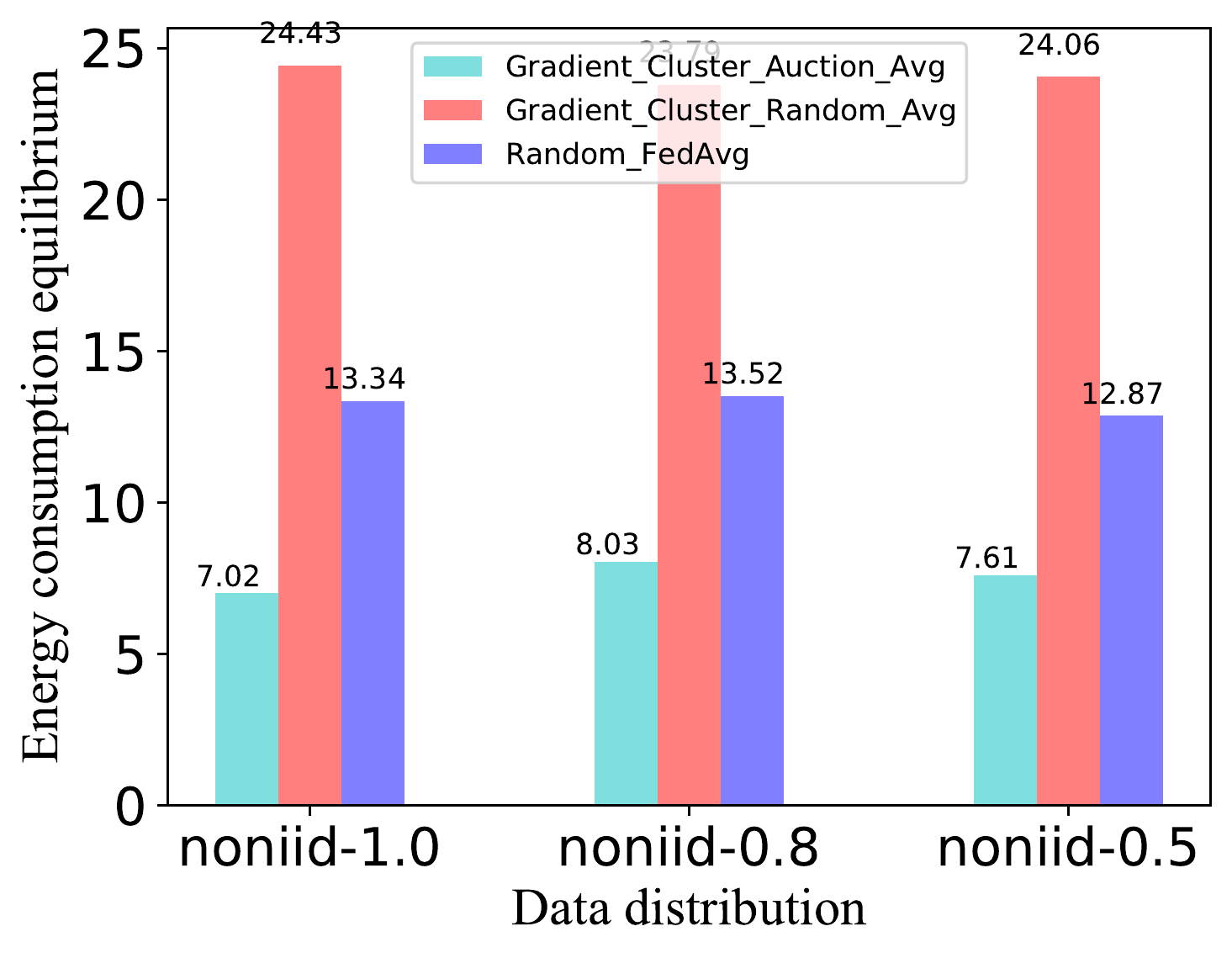}}

\caption{ The Energy balance on different levels of Non-IID under 100 clients, when using Avg. }
\label{fig10}
\end{figure*}

\begin{figure*}[!t]
\centering
\subfigure[Fashion MNIST]{\includegraphics[width=2.3in,angle=0]{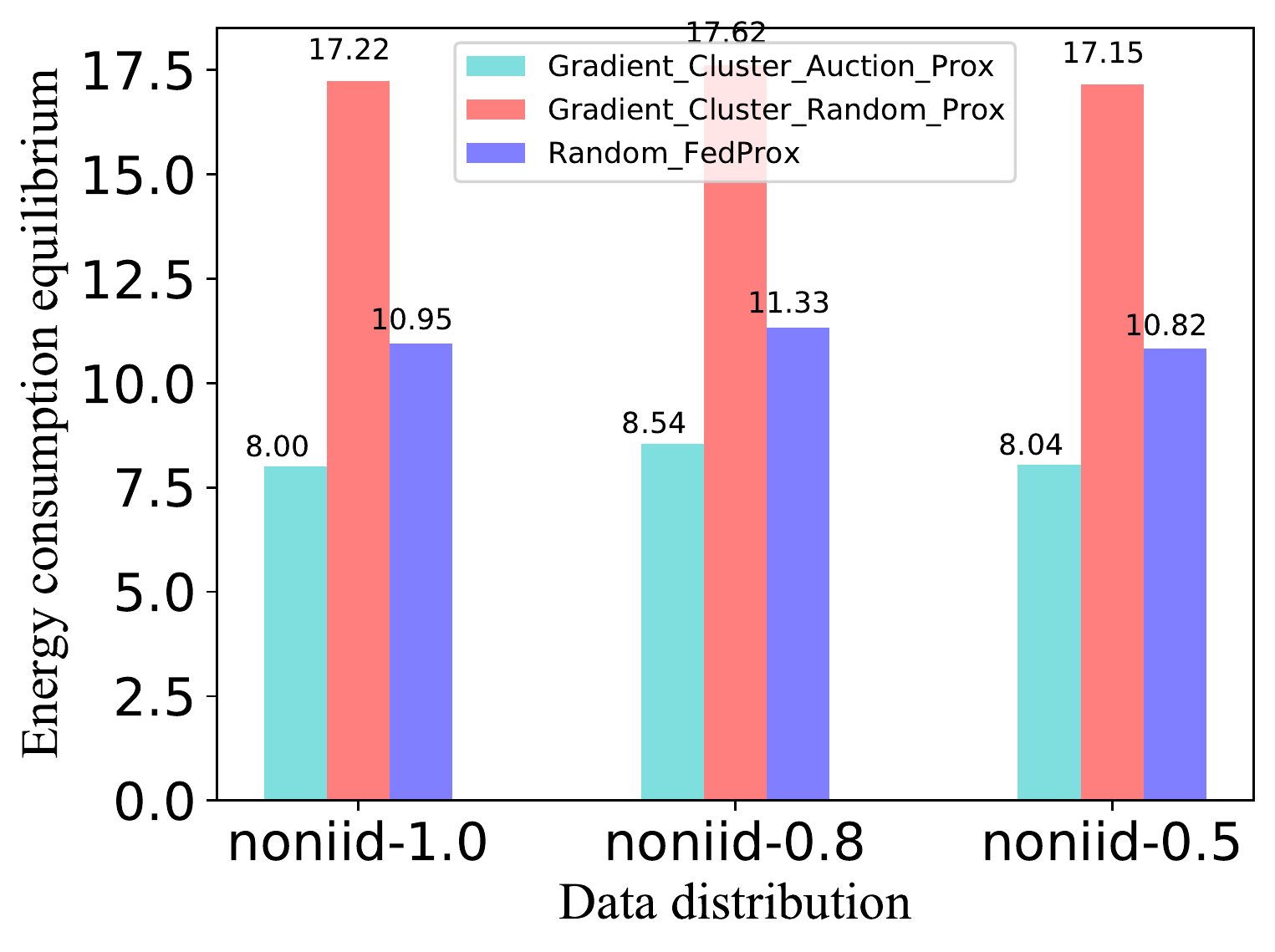}}
\subfigure[MNIST]{\includegraphics[width=2.2in,angle=0]{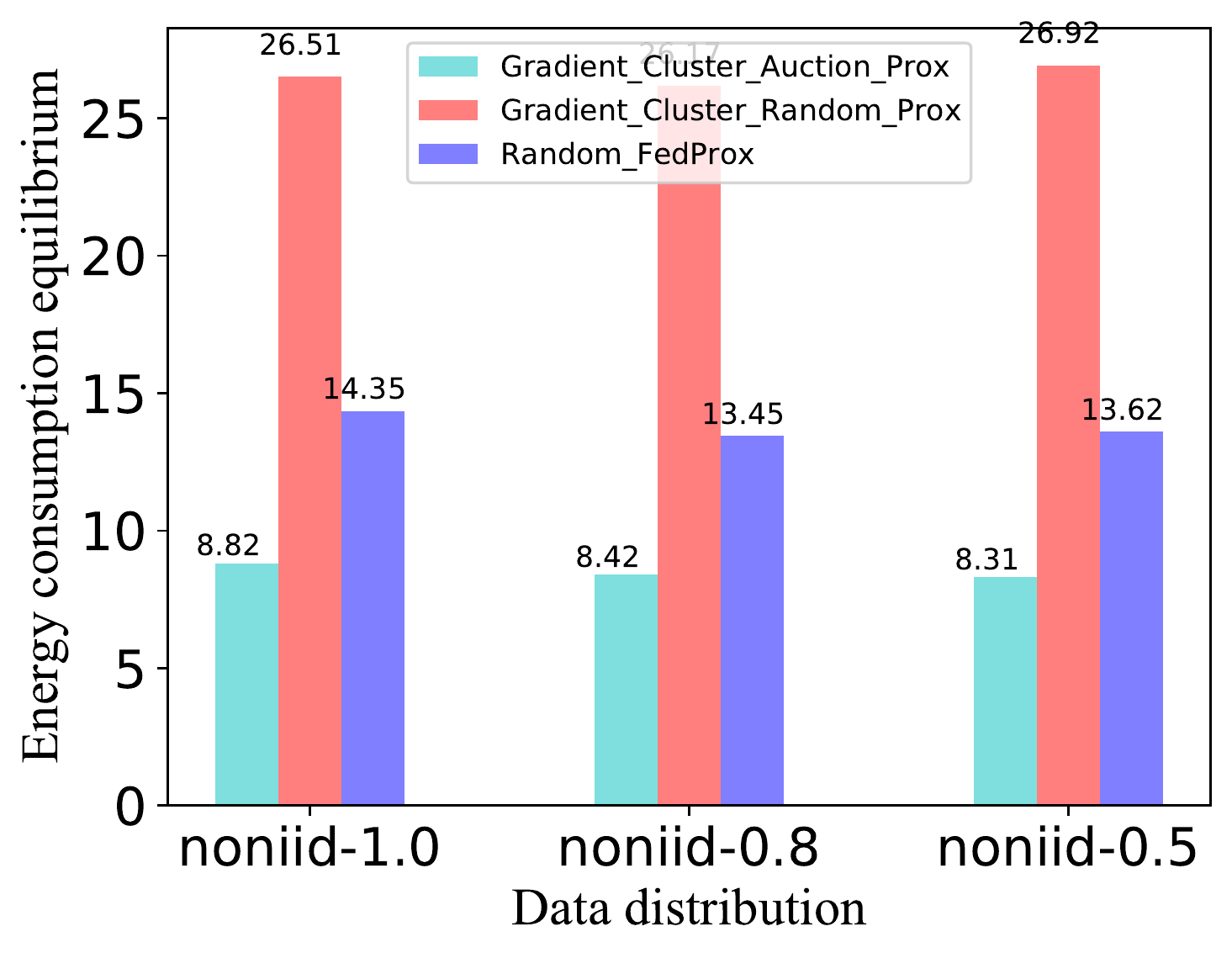}}
\subfigure[CIFAR-10]{\includegraphics[width=2.2in,angle=0]{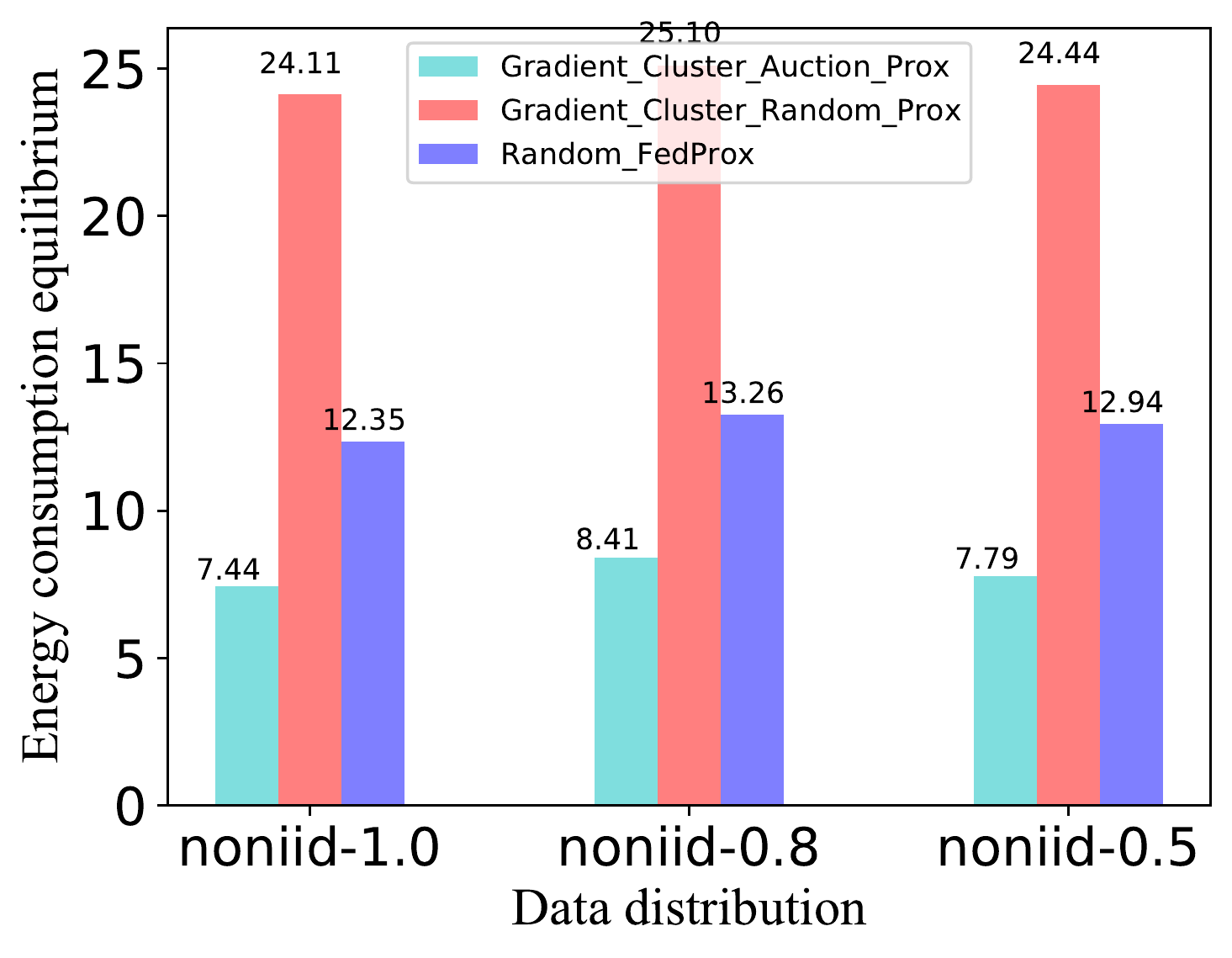}}

\caption{ The Energy balance on different levels of Non-IID under 100 clients, when using Prox. }
\label{fig11}
\end{figure*}

\subsubsection{Energy consumption}
In terms of energy consumption, for the balance of remaining energy, we use the standard deviation of all clients' remaining energy in the system as the metric to measure the balance of energy consumption. The system energy consumption is balanced when the standard deviation is smaller. We set the system client to 100 and then verify the scheme's effectiveness in three different Non-IID scenarios. As shown in Fig. 9 and Fig. 10, our proposed auction-based selection scheme shows a better balance than the other schemes. The reason is that the auction-based client's selection scheme fully considers the clients' remaining energy consumption in the cost function and then bids according to the optimal solution that satisfies the Nash equilibrium to achieve the balance of energy consumption in the system. Since the FedAvg and FedProx scheme uses a random sample selection strategy, some clients with many local samples may consume too much energy.

\section{Conclusion}
 In our research, we are committed to solving the data heterogeneity problem faced by the federated learning system by designing a client selection scheme. Thus, we design a client selection method based on initial gradient clustering. This method mainly includes three innovations. Firstly, we introduce the concept of a federated virtual data set for the first time to provide a solution to the data heterogeneity of the federated learning system. Secondly, we proposed the sample windows mechanism  , whose goal is to alleviate the impact of local data imbalance on clustering accuracy. Thirdly, we prove that our proposed client's selection method can converge to the approximate optimal solution under the stochastic gradient descent algorithm.

Furthermore, we designed an auction-based client selection algorithm in each cluster, which aims to solve the imbalance of resource consumption caused by the random selection of clients. At the same time, we introduced energy-balanced metrics in the federal learning scenario for the first time. Simulation results show that our proposed client selection scheme can achieve better performance in the AI model's convergence rate and the energy consumption balance of the system.

\appendices
\section{proof of theorem1}
In this section, we give assumptions, lemmas, and proofs related to Theorem 1.

% you can choose not to have a title for an appendix
% if you want by leaving the argument blank
\subsection{Assumption}
Firstly, we make some assumptions about the objective loss function $f(*)$.

\noindent\textbf{Assumption 1} The objective function $f(*)$ is a $L$-smooth function, satisfying:$$f\left ( y\right )-f\left ( x\right )  \le  \nabla f\left ( x\right )^{T}\left ( y-x\right )+\frac{1}{2}\times L\left \| y-x\right \|_{2}^{2}$$

\noindent\textbf{Assumption 2} The objective function  is a $\beta$ strongly convex function, satisfying:
$$f\left ( y\right )-f\left ( x\right ) >  \nabla f\left ( x\right )^T\left ( y-x\right )+\frac{1}{2}\beta \left \| y-x\right \|_{2}^{2}.$$

Gaurush et al.\cite{DBLP:conf/ijcnn/BriggsFA21} make the following assumptions which give the expected bounds of the sample gradient, as show in the follow assumptions.

\noindent\textbf{Assumption3} Bounds of gradient expectation, $\exists \mu,\mu_G(0<\mu\le\mu_G)$ the expectations for random sampling: $$\mu\left \| \nabla f \left ( w_t,D\right ) \right \|<E_{\xi_t \in D}\left[g(w_t,\xi_t)\right ]<\mu_G\left \|\nabla f\left ( w_t,D\right ) \right \|$$
where $\nabla f\left ( w_t,D\right )$ represents the gradient mean of all samples, $D$ is the set of all local samples.

\noindent\textbf{Assumption4} Bounds of the second-order norm of the gradient, $\exists M,M_G(0<M\le M_G)$, the expectations for random sampling are met:
$$E_{\xi_t}\left [ \left \| g(w_t,\xi_t)\right \|^{2}\right ] \le M+M_G\left \|\nabla f \right \|^{2}$$
where $\nabla f$ represents the gradient mean of all samples.

In addition, we assume that the data distribution in different clusters also meets assumption 3 and assumption 4. It is obviously that the corresponding factor $\mu$, $\mu_G$, $M$, $M_G$ of different clusters are different in the gradient-based clustering scheme. And,we use $\mu_j$, $\mu_{Gj}$, $M_j$, $M_{Gj}$ to represent the corresponding parameters in cluster $j$.

\subsection{Lemma}
\noindent\textbf{Lemma 1} The expected bounds of the gradient of the cluster sampling strategies are:
$$\mu_{E}\left \|\nabla f \left ( w_t,D\right ) \right \|\le E_{\xi _t}\left [ g\left ( w_t,\xi _t\right )\right ]\le \mu_{GE}\left \|\nabla f \left ( w_t,D\right ) \right \|$$
where $J$ is the total number of clusters, and
$\mu_E=\frac{1}{J}\sum _{j=1}^{J}\mu_j $, $\mu_{GE}=\frac{1}{J}\sum _{j=1}^{J}\mu_{Gj}.\\$

\textbf{\emph{Proof:}}

According to \textbf{Assumption 3}, each cluster satisfies:
$$\mu_j\left \| \nabla f \left ( w_t,D\right ) \right \|<E_{\xi_t \in D}\left[g(w_t,\xi_t)\right ]<\mu_{Gj}\left \|\nabla f\left ( w_t,D\right ) \right \|.$$
Summation on both sides of inequality:
$$\frac{1}{J}\sum _{j=1}^{J}\mu_j\left \| \nabla f \left ( w_t,D\right ) \right \|  \le \frac{1}{J}\sum _{j=1}^{J}E_{\zeta \in D}\left[g(w_t,\xi_t)\right ]$$
$$\frac{1}{J}\sum _{j=1}^{J}E_{\zeta \in D}\left[g(w_t,\xi_t)\right ] \le \frac{1}{J}\sum _{j=1}^{J}\mu_{Gj}\left \|\nabla f\left ( w_t,D\right ) \right \|$$
then
$$\mu_E=\frac{1}{J}\sum _{j=1}^{J}\mu_j, \mu_{GE}=\frac{1}{J}\sum _{j=1}^{J}\mu_{Gj}$$
$$\mu_{E}\left \|\nabla f \left ( w_t,D\right ) \right \|\le E_{\xi _t}\left [ g\left ( w_t,\xi _t\right )\right ]\le \mu_{GE}\left \|\nabla f \left ( w_t,D\right ) \right \|$$

\noindent\textbf{Lemma 2} The expected bounds of the gradient of the cluster sampling strategies are:
$$E_{\xi_t}\left [ \left \| g(w_t,\xi_t)\right \|^{2}\right ] \le M+M_{GE}\left \|\nabla f \right \|^{2}$$
where $M_{GE}=\frac{1}{J}\sum _{j=1}^{J}M_{Gj}.\\$

\textbf{\emph{Proof:}}

According to \textbf{Assumption 4}, each cluster satisfies:
\begin{align*}
 & E_{\xi_t}\left [ \left \| g(w_t,\xi_t)\right \|^{2}\right ] \le M+M_{Gj}\left \|\nabla f \right \|^{2}.
\end{align*}
Summation on both sides of inequality:
\begin{align*}
\frac{1}{J}\sum _{j=1}^{J}E_{\xi_t}\left [ \left \| g(w_t,\xi_t)\right \|^{2}\right ]&\le \frac{1}{J}\sum _{j=1}^{J}\left (M+M_{Gj}\left \|\nabla f \right \|^{2} \right )\\
&\le M+\frac{1}{J}\sum _{j=1}^{J}\left (M_{Gj}\left \|\nabla f \right \|^{2} \right )\\
& \le M+M_{GE}\left \|\nabla f \right \|^{2}
\end{align*}
then
$$M_{GE}=\frac{1}{J}\sum _{j=1}^{J}M_{Gj}$$
$$E_{\xi_t}\left [ \left \| g(w_t,\xi_t)\right \|^{2}\right ] \le M+M_{GE}\left \|\nabla f \right \|^{2}.$$

\noindent\textbf{Lemma 3} Under the assumption of $L$ smooth function,
\begin{multline*}
E\left [ f(w_{t+1})\right ]-f\left ( w_t\right )\le-\theta I\eta _t\nabla f \left ( w_t\right ) E_{\xi _t}\left [ g\left ( w_t,\xi_t\right )\right ]\\
+ \frac{1}{2}\theta^2I^2\eta _t^2LE_{\xi _t}\left [ \left \| g\left ( w_t,\xi_t\right )\right \|^2\right ]
\end{multline*}

\textbf{\emph{Proof:}}

According to \textbf{Assumption 1},
\begin{align*}
f(w_{t+1})-f(w_{t})&\le\nabla f\left (w_{t} \right )^{T}(w_{t+1}-w_t)+\frac{L}{2}\left \|w_{t+1}-w_t \right \|^2 \\
&=\nabla f\left (w_{t} \right )^{T}\eta _t\sum_{k=1}^{K}p_k\sum_{i=0}^{I-1}\nabla f\left (w_{t,i},x_{t,i}^{k} \right )\\
&\quad+\frac{1}{2}\eta _t^2L\left \|\sum_{k=1}^{K}p_k\sum_{i=0}^{I-1}\nabla f\left (w_{t,i},x_{t,i}^{k} \right ) \right \|^2
\end{align*}

Take expectations on both sides of the inequality, we can conclude:
\begin{multline*}
E\left [ f(w_{t+1})\right ]-f\left ( w_t\right )\le-\theta I\eta _t\nabla f \left ( w_t\right ) E_{\xi _t}\left [ \left \| g\left ( w_t,\xi_t\right )\right \|^2\right ]\\+ \frac{1}{2}\theta^2I^2\eta _t^2LE_{\xi _t}\left [ g\left ( w_t,\xi_t\right )\right ]
\end{multline*}

\noindent\textbf{Lemma4} Under the assumption of 1, 3, 4,
\begin{multline*}
E\left [ f(w_{t+1})\right ]-f\left ( w_t\right )\le- \theta I\eta _t\left ( \mu-\frac{\eta_t}{2} \theta ILM_G\right ) \left \|\nabla f \left ( w_t\right ) \right \|^2\\+\frac{\eta_t^2}{2}\theta^2I^2LM
\end{multline*}

\textbf{\emph{Proof:}}

From \textbf{lemma 3}, it can be concluded that:
\begin{multline*}
 E\left [ f(w_{t+1})\right ]-f\left ( w_t\right )\le-\theta I\eta _t\nabla f \left ( w_t\right ) E_{\xi _t}\left [ \left \| g\left ( w_t,\xi_t\right )\right \|^2\right ]\\+ \frac{1}{2}\theta^2I^2\eta _t^2LE_{\xi _t}\left [ g\left ( w_t,\xi_t\right )\right ].
\end{multline*}

\noindent According to \textbf{Assumption 3:}
$$ \nabla f \left ( w_t\right ) ^{T}E_{\xi _t}\left [ g\left ( w_t,\xi_t\right )\right ]\le \mu\left \|\nabla f \left ( w_t\right ) \right \|^2$$

\noindent then
$$E\left [ f(w_{t+1})\right ]-f\left ( w_t\right )\le-\theta I\eta _t \left \|\nabla f \left ( w_t\right ) \right \|^2$$
$$\qquad\qquad\quad\qquad\qquad\qquad\qquad+ \frac{1}{2}\theta^2I^2\eta _t^2LE_{\xi _t}\left [ \left \| g\left ( w_t,\xi_t\right )\right \|^2\right ].$$
\noindent According to \textbf{Assumption 4}:
$$E_{\xi _t}\left [ \left \| g\left ( w_t,\xi_t\right )\right \|^2\right]\le\mu_G\left \|\nabla f \left ( w_t\right )  \right \|^2$$
\noindent then:
$$E\left [ f(w_{t+1})\right ]-f\left ( w_t\right )\le-\theta I\eta _t\mu\left \|\nabla f \left ( w_t\right )\right \|^2$$
$$+ \frac{L}{2}\theta^2I^2\eta _t^2(M+M_G)\left \|\nabla f \left ( w_t\right )  \right \|^2$$
$$\le -\theta I\eta _t(\mu - \frac{\eta _t}{2}ILM_G)\left \|\nabla f \left ( w_t\right )  \right \|^2$$
$$+\frac{\theta^2I^2\eta _t^2}{2}LM$$

\noindent\textbf{Lemma5} Under the \textbf{Assumption 1}, and $w^*$ is the optimal model parameter,
$$\left \|w_t-w^* \right \|^2\ge \frac{2}{L}(f(w_t)-f(w^*)).$$

\noindent\textbf{Lemma6} Under the \textbf{Assumption 2}, and $w^*$ is the optimal model parameter,
$$\left \| w_t-w^*\right \|\le\frac{2}{\beta }\left \|\nabla f \left ( w_t\right )  \right \|.$$
Gaurush et al.\cite{DBLP:conf/ijcnn/BriggsFA21} gave proofs of Lemma 5 and Lemma 6.

\noindent\textbf{Lemma7} Under the \textbf{Assumption 1}, \textbf{Assumption 2}, and $w^*$ is the optimal model parameter,
$$\left \|\nabla f \left ( w_t\right )  \right\|^2\ge\frac{\beta^2}{2L}(f(w_t)-f(w^*))$$

\textbf{\emph{Proof:}}

According to \textbf{lemma 5}:
$$\left \|w_t-w^* \right \|^2\ge \frac{2}{L}(f(w_t)-f(w^*)).$$

According to \textbf{lemma 6}::
$$\left \| w_t-w^*\right \|\le\frac{2}{\beta }\left \|\nabla f \left ( w_t\right )  \right \|$$
$$\left \|w^*-w_t \right \|^2\le\frac{4}{\beta^2 }\left \|\nabla f \left ( w_t\right )  \right \|^2$$
$$\frac{4}{\beta^2 }\left \|\nabla f \left ( w_t\right )  \right \|^2\ge\left \|w^*-w_t \right \|^2\ge\frac{2}{L}(f(w_t)-f(w^*))$$
$$\left \|\nabla f \left ( w_t\right )  \right\|^2\ge\frac{\beta^2}{2L}(f(w_t)-f(w^*))$$

\subsection{Theorem}
\textbf{Theorem 1} Under the assumption 1 to 4, adopting clustering clients sampling strategy while fixing the step size $\eta_t=\eta$ satisfies:
$$E\left [ f(w_{t+1})\right ]-f\left ( w_t\right )\le(1-B_1)^{t-1}(f(w_1)-f(w^*)-A_1)+A_1$$
where $0<\eta\le \frac{\mu_E}{LM_G}$, $A_1=\frac{2\eta IL^2M}{\mu_E \beta^2}$, $B_1=\frac{\theta I\eta\mu_E\beta^2}{4L}.\\$

\textbf{\emph{Proof:}}

According to \textbf{lemma 4}:
\begin{multline*}
E\left [ f(w_{t+1})\right ]-f\left ( w_t\right )\le - \theta I\eta\left ( \mu-\frac{\eta}{2}ILM_G\right ) \left \|\nabla f \left ( w_t\right ) \right \|^2\\
+\frac{\eta^2}{2}\theta^2I^2LM
\end{multline*}
$$\because 0<\eta\le \frac{\mu_E}{LM_G}$$
$$\therefore -\theta I\eta\left ( \mu-\frac{\eta}{2}\theta ILM_G\right )\le -\frac{\mu}{2}\theta I\eta$$
$$\therefore E\left [ f(w_{t+1})\right ]-f\left ( w_t\right )\le -\frac{\mu}{2}\theta I\eta \left \|\nabla f \left ( w_t\right ) \right \|^2+\frac{\eta^2}{2}\theta^2I^2LM$$
$$E\left [ f(w_{t+1})\right ]\le f\left ( w_t\right ) -\frac{\mu}{2}\theta I\eta \left \|\nabla f \left ( w_t\right ) \right \|^2+\frac{\eta^2}{2}\theta I^2LM.$$
Subtract $f(w^*)$ from both sides of the inequality:
\begin{multline*}
E\left [ f(w_{t+1})\right ]-f(w^*)\le f\left ( w_t\right )-f(w^*) -\frac{\mu}{2}\theta I\eta \left \|\nabla f \left ( w_t\right ) \right \|^2\\ +\frac{\eta^2}{2}\theta^2I^2LM.
\end{multline*}
According to \textbf{lemma 7}:,
$$\left \|\nabla f \left ( w_t\right )  \right\|^2\ge\frac{\beta^2}{2L}(f(w_t)-f(w^*)).$$
Multiply both sides of the inequality by $-\frac{\mu}{2}\theta I\eta$
$$-\frac{\mu}{2}\theta I\eta\left \|\nabla f \left ( w_t\right )  \right\|^2\le-\frac{\theta I\eta\mu\beta^2}{4L}(f(w_t)-f(w^*))$$
\begin{multline*}
E\left [ f(w_{t+1})\right ]-f(w^*)\le (1-\frac{\theta I\eta\mu\beta^2}{4L})(f(w_t)-f(w^*))\\
+\frac{\eta^2}{2}\theta^2I^2LM.
\end{multline*}
Both sides of the inequality are subtracted by the introduced item $A$, then
\begin{multline*}
E\left [ f(w_{t+1})\right ]-f(w^*)-A\le (1-\frac{\theta I\eta\mu\beta^2}{4L})(f(w_t)-f(w^*))\\
+\frac{\eta^2}{2}\theta^2I^2LM-A.
\end{multline*}
Suppose the following equation holds
\begin{multline*}
(1-\frac{\theta I\eta\mu\beta^2}{4L})(f(w_t)-f(w^*))+\frac{\eta^2}{2}\theta^2 I^2LM-A\\
=(1-\frac{\theta I\eta\mu\beta^2}{4L})(f(w_t)-f(w^*)-A).
\end{multline*}
It can be deduced that
$$A=\frac{2\eta\theta  IL^2M}{\mu \beta ^2}$$
then:
\begin{multline*}
E\left [ f(w_{t+1})\right ]-f(w^*)-\frac{2\eta\theta  IL^2M}{\mu \beta ^2}\\
\le(1-\frac{I\eta\theta \mu\beta^2}{4L})(f(w_t)-f(w^*)-\frac{2\eta\theta  IL^2M}{\mu \beta ^2})\\
\qquad\le(1-\frac{I\eta\theta \mu\beta^2}{4L})^{2}(f(w_{t-1})-f(w^*)-\frac{2\eta\theta  IL^2M}{\mu \beta ^2})\\
\le(1-\frac{\theta I\eta\mu\beta^2}{4L})^{t-1}(f(w_1)-f(w^*)-\frac{2\eta\theta  IL^2M}{\mu \beta ^2})
\end{multline*}

$$E\left [ f(w_{t+1})\right ]-f(w^*)
\le(1-B)^{t-1}(f(w_1)-f(w^*)-A)+A$$
where $B=\frac{I\eta\theta \mu\beta^2}{4L}$,
when using clustering clients sampling strategy, $\mu=\mu_E$, then:
\begin{multline*}
E\left [ f(w_{t+1})\right ]-f(w^*)\le\\
(1-\frac{\theta I\eta\mu_E\beta^2}{4L})^{t-1}(f(w_1)-f(w^*)-\frac{2\eta\theta  IL^2M}{\mu_E \beta ^2})+\frac{2\eta \theta IL^2M}{\mu_E \beta ^2}.
\end{multline*}

\bibliographystyle{IEEEtran}
\bibliography{bare_jrnl}

% Generated by IEEEtran.bst, version: 1.14 (2015/08/26)
\begin{thebibliography}{10}
\providecommand{\url}[1]{#1}
\csname url@samestyle\endcsname
\providecommand{\newblock}{\relax}
\providecommand{\bibinfo}[2]{#2}
\providecommand{\BIBentrySTDinterwordspacing}{\spaceskip=0pt\relax}
\providecommand{\BIBentryALTinterwordstretchfactor}{4}
\providecommand{\BIBentryALTinterwordspacing}{\spaceskip=\fontdimen2\font plus
\BIBentryALTinterwordstretchfactor\fontdimen3\font minus
  \fontdimen4\font\relax}
\providecommand{\BIBforeignlanguage}[2]{{%
\expandafter\ifx\csname l@#1\endcsname\relax
\typeout{** WARNING: IEEEtran.bst: No hyphenation pattern has been}%
\typeout{** loaded for the language `#1'. Using the pattern for}%
\typeout{** the default language instead.}%
\else
\language=\csname l@#1\endcsname
\fi
#2}}
\providecommand{\BIBdecl}{\relax}
\BIBdecl

\bibitem{Mcmahan}
B.~McMahan, E.~Moore, D.~Ramage, S.~Hampson, and B.~A. y~Arcas,
  ``Communication-efficient learning of deep networks from decentralized
  data,'' in \emph{Artificial Intelligence and Statistics}.\hskip 1em plus
  0.5em minus 0.4em\relax PMLR, 2017, pp. 1273--1282.

\bibitem{Wang}
H.~Wang, Z.~Kaplan, D.~Niu, and B.~Li, ``Optimizing federated learning on
  non-iid data with reinforcement learning,'' in \emph{IEEE INFOCOM 2020-IEEE
  Conference on Computer Communications}.\hskip 1em plus 0.5em minus
  0.4em\relax IEEE, 2020, pp. 1698--1707.

\bibitem{Sattler}
F.~Sattler, K.-R. M{\"u}ller, T.~Wiegand, and W.~Samek, ``On the byzantine
  robustness of clustered federated learning,'' in \emph{ICASSP 2020-2020 IEEE
  International Conference on Acoustics, Speech and Signal Processing
  (ICASSP)}.\hskip 1em plus 0.5em minus 0.4em\relax IEEE, 2020, pp. 8861--8865.

\bibitem{Khan}
L.~U. Khan, M.~Alsenwi, Z.~Han, and C.~S. Hong, ``Self organizing federated
  learning over wireless networks: A socially aware clustering approach,'' in
  \emph{2020 International Conference on Information Networking (ICOIN)}.\hskip
  1em plus 0.5em minus 0.4em\relax IEEE, 2020, pp. 453--458.

\bibitem{Jiang}
R.~Jiang and S.~Zhou, ``Cluster-based cooperative digital over-the-air
  aggregation for wireless federated edge learning,'' in \emph{2020 IEEE/CIC
  International Conference on Communications in China (ICCC)}.\hskip 1em plus
  0.5em minus 0.4em\relax IEEE, 2020, pp. 887--892.

\bibitem{SattlerF}
F.~Sattler, K.-R. M{\"u}ller, and W.~Samek, ``Clustered federated learning:
  Model-agnostic distributed multitask optimization under privacy
  constraints,'' \emph{IEEE Transactions on Neural Networks and Learning
  Systems}, 2020.

\bibitem{GhoshA}
A.~Ghosh, J.~Chung, D.~Yin, and K.~Ramchandran, ``An efficient framework for
  clustered federated learning,'' \emph{arXiv preprint arXiv:2006.04088}, 2020.

\bibitem{DBLP:conf/icc/CaiLZY20}
L.~Cai, D.~Lin, J.~Zhang, and S.~Yu, ``Dynamic sample selection for federated
  learning with heterogeneous data in fog computing,'' in \emph{ICC 2020-2020
  IEEE International Conference on Communications (ICC)}.\hskip 1em plus 0.5em
  minus 0.4em\relax IEEE, 2020, pp. 1--6.

\bibitem{khan2020federated}
L.~U. Khan, S.~R. Pandey, N.~H. Tran, W.~Saad, Z.~Han, M.~N. Nguyen, and C.~S.
  Hong, ``Federated learning for edge networks: Resource optimization and
  incentive mechanism,'' \emph{IEEE Communications Magazine}, vol.~58, no.~10,
  pp. 88--93, 2020.

\bibitem{le2020auction}
T.~H.~T. Le, N.~H. Tran, Y.~K. Tun, Z.~Han, and C.~S. Hong, ``Auction based
  incentive design for efficient federated learning in cellular wireless
  networks,'' in \emph{2020 IEEE Wireless Communications and Networking
  Conference (WCNC)}.\hskip 1em plus 0.5em minus 0.4em\relax IEEE, 2020, pp.
  1--6.

\bibitem{DBLP:conf/icml/BhagojiCMC19}
A.~N. Bhagoji, S.~Chakraborty, P.~Mittal, and S.~Calo, ``Analyzing federated
  learning through an adversarial lens,'' in \emph{International Conference on
  Machine Learning}.\hskip 1em plus 0.5em minus 0.4em\relax PMLR, 2019, pp.
  634--643.

\bibitem{DBLP:conf/iclr/XieHCL20}
C.~Xie, K.~Huang, P.-Y. Chen, and B.~Li, ``Dba: Distributed backdoor attacks
  against federated learning,'' in \emph{International Conference on Learning
  Representations}, 2019.

\bibitem{DBLP:conf/infocom/WangSZSWQ19}
Z.~Wang, M.~Song, Z.~Zhang, Y.~Song, Q.~Wang, and H.~Qi, ``Beyond inferring
  class representatives: User-level privacy leakage from federated learning,''
  in \emph{IEEE INFOCOM 2019-IEEE Conference on Computer Communications}.\hskip
  1em plus 0.5em minus 0.4em\relax IEEE, 2019, pp. 2512--2520.

\bibitem{DBLP:conf/esorics/TolpeginTGL20}
V.~Tolpegin, S.~Truex, M.~E. Gursoy, and L.~Liu, ``Data poisoning attacks
  against federated learning systems,'' in \emph{European Symposium on Research
  in Computer Security}.\hskip 1em plus 0.5em minus 0.4em\relax Springer, 2020,
  pp. 480--501.

\bibitem{DBLP:journals/corr/abs-1911-12560}
J.~Lin, M.~Du, and J.~Liu, ``Free-riders in federated learning: Attacks and
  defenses,'' \emph{arXiv preprint arXiv:1911.12560}, 2019.

\bibitem{DBLP:conf/vcip/0003HS18}
X.~Yao, C.~Huang, and L.~Sun, ``Two-stream federated learning: Reduce the
  communication costs,'' in \emph{2018 IEEE Visual Communications and Image
  Processing (VCIP)}.\hskip 1em plus 0.5em minus 0.4em\relax IEEE, 2018, pp.
  1--4.

\bibitem{DBLP:journals/corr/KonecnyMYRSB16}
J.~Kone{\v{c}}n{\`y}, H.~B. McMahan, F.~X. Yu, P.~Richt{\'a}rik, A.~T. Suresh,
  and D.~Bacon, ``Federated learning: Strategies for improving communication
  efficiency,'' \emph{arXiv preprint arXiv:1610.05492}, 2016.

\bibitem{DBLP:conf/dcc/LiH19}
H.~Li and T.~Han, ``An end-to-end encrypted neural network for gradient updates
  transmission in federated learning,'' \emph{arXiv preprint arXiv:1908.08340},
  2019.

\bibitem{DBLP:journals/tnn/ZhuJ20}
H.~Zhu and Y.~Jin, ``Multi-objective evolutionary federated learning,''
  \emph{IEEE Transactions on Neural Networks and Learning Systems}, vol.~31,
  no.~4, pp. 1310--1322, 2019.

\bibitem{DBLP:conf/icca/LiFL20}
L.~Li, Y.~Fan, and K.~Lin, ``A survey on federated learning,'' in \emph{16th
  {IEEE} International Conference on Control {\&} Automation, {ICCA} 2020,
  Singapore, October 9-11, 2020}.\hskip 1em plus 0.5em minus 0.4em\relax
  {IEEE}, 2020, pp. 791--796.

\bibitem{DBLP:conf/iclr/LiHYWZ20}
X.~Li, K.~Huang, W.~Yang, S.~Wang, and Z.~Zhang, ``On the convergence of fedavg
  on non-iid data,'' \emph{arXiv preprint arXiv:1907.02189}, 2019.

\bibitem{DBLP:journals/corr/abs-1806-00582}
Y.~Zhao, M.~Li, L.~Lai, N.~Suda, D.~Civin, and V.~Chandra, ``Federated learning
  with non-iid data,'' \emph{arXiv preprint arXiv:1806.00582}, 2018.

\bibitem{DBLP:journals/jsac/WangTSLMHC19}
S.~Wang, T.~Tuor, T.~Salonidis, K.~K. Leung, C.~Makaya, T.~He, and K.~Chan,
  ``Adaptive federated learning in resource constrained edge computing
  systems,'' \emph{IEEE Journal on Selected Areas in Communications}, vol.~37,
  no.~6, pp. 1205--1221, 2019.

\bibitem{DBLP:conf/mlsys/LiSZSTS20}
T.~Li, A.~K. Sahu, M.~Zaheer, M.~Sanjabi, A.~Talwalkar, and V.~Smith,
  ``Federated optimization in heterogeneous networks,'' \emph{arXiv preprint
  arXiv:1812.06127}, 2018.

\bibitem{DBLP:journals/fgcs/MothukuriPPHDS21}
V.~Mothukuri, R.~M. Parizi, S.~Pouriyeh, Y.~Huang, A.~Dehghantanha, and
  G.~Srivastava, ``A survey on security and privacy of federated learning,''
  \emph{Future Generation Computer Systems}, vol. 115, pp. 619--640, 2020.

\bibitem{DBLP:conf/iclr/WangYSPK20}
H.~Wang, M.~Yurochkin, Y.~Sun, D.~Papailiopoulos, and Y.~Khazaeni, ``Federated
  learning with matched averaging,'' \emph{arXiv preprint arXiv:2002.06440},
  2020.

\bibitem{DBLP:conf/ijcnn/BriggsFA20}
C.~Briggs, Z.~Fan, and P.~Andras, ``Federated learning with hierarchical
  clustering of local updates to improve training on non-iid data,''
  \emph{arXiv preprint arXiv:2004.11791}, 2020.

\bibitem{DBLP:conf/ijcnn/BriggsFA21}
\BIBentryALTinterwordspacing
G.~Hiranandani and P.~Chiu, ``Variations of the stochastic gradient descent for
  multi-label classication loss functions,'' \emph{Online}, 2020. [Online].
  Available: \url{https://gaurush.com/assets/docs/ece_566.pdf}
\BIBentrySTDinterwordspacing

\end{thebibliography}

\ifCLASSOPTIONcaptionsoff
  \newpage
\fi

\end{document}